\documentclass[twoside,twocolumn,10pt]{article}
\usepackage{graphicx,wrapfig}

\def\,{\mskip 3mu} \def\>{\mskip 4mu plus 2mu minus 4mu} \def\;{\mskip 5mu plus 5mu} \def\!{\mskip-3mu}
\def\dispmuskip{\thinmuskip= 3mu plus 0mu minus 2mu \medmuskip=  4mu plus 2mu minus 2mu \thickmuskip=5mu plus 5mu minus 2mu}
\def\textmuskip{\thinmuskip= 0mu                    \medmuskip=  1mu plus 1mu minus 1mu \thickmuskip=2mu plus 3mu minus 1mu}
\textmuskip
\def\beq{\dispmuskip\begin{equation}}    \def\eeq{\end{equation}\textmuskip}
\def\beqn{\dispmuskip\begin{displaymath}}\def\eeqn{\end{displaymath}\textmuskip}
\def\bqa{\dispmuskip\begin{eqnarray}}    \def\eqa{\end{eqnarray}\textmuskip}
\def\bqan{\dispmuskip\begin{eqnarray*}}  \def\eqan{\end{eqnarray*}\textmuskip}

\newenvironment{keywords}{\vskip 3ex\noindent {\bf\Large Keywords}\vskip 2ex\noindent}{\par\vskip 1ex}
\def\subsection#1{\vspace{1ex plus 0.5ex minus 0.5ex}\noindent{\bfseries\boldmath{#1.}}}

\def\odt{{\textstyle{1\over 2}}}
\def\eps{\varepsilon}

\def\approxleq{\mbox{\raisebox{-0.8ex}{$\stackrel{\displaystyle<}\sim$}}} 
\def\approxgeq{\mbox{\raisebox{-0.8ex}{$\stackrel{\displaystyle>}\sim$}}} 
\def\SetR{{I\!\!R}}

\begin{document}

\title{\vskip -10mm\normalsize\sc Technical Report \hfill IDSIA-16-06
\vskip 2mm\bf\huge\hrule height5pt \vskip 6mm
Fitness Uniform Optimization
\vskip 6mm \hrule height2pt \vskip 5mm}
\author{
{\bf Marcus Hutter}\\[2mm]
IDSIA, Galleria 2, CH-6928\\ Manno-Lugano, Switzerland\\
marcus@idsia.ch
\and
{\bf Shane Legg}\\[2mm]
IDSIA, Galleria 2, CH-6928\\ Manno-Lugano, Switzerland\\
shane@idsia.ch}

\maketitle

\begin{abstract}
In evolutionary algorithms, the fitness of a population increases with
time by mutating and recombining individuals and by a biased selection
of more fit individuals. The right selection pressure is critical in
ensuring sufficient optimization progress on the one hand and in
preserving genetic diversity to be able to escape from local optima on
the other hand. Motivated by a universal similarity relation on the
individuals, we propose a new selection scheme, which is uniform in
the fitness values. It generates selection pressure toward sparsely
populated fitness regions, not necessarily toward higher fitness, as
is the case for all other selection schemes. We show analytically on a
simple example that the new selection scheme can be much more
effective than standard selection schemes.  We also propose a new
deletion scheme which achieves a similar result via deletion and show
how such a scheme preserves genetic diversity more effectively than
standard approaches.  We compare the performance of the new schemes to
tournament selection and random deletion on an artificial deceptive
problem and a range of NP-hard problems: traveling salesman, set
covering and satisfiability.
\end{abstract}


\section{Introduction}\label{secInt}

\subsection{Evolutionary algorithms (EA)}
Evolutionary algorithms are capable of solving complicated
optimization tasks in which an objective function $f:I\to\SetR$ shall
be maximized. $i\in I$ is an individual from the set $I$ of feasible
solutions. Infeasible solutions due to constraints may also be
considered by reducing $f$ for each violated constraint. A population
$P$ is a multi-set of individuals from $I$ which is maintained and
updated as follows: one or more individuals are selected according to
some selection strategy.
In generation based EAs, the selected individuals are recombined
(e.g.\ crossover) and mutated, and constitute the new population.
We prefer the more incremental, steady-state population update,
which selects (and possibly deletes) only one or two individuals from
the current population and adds the newly recombined and mutated
individuals to it.
We are interested in finding a single individual of maximal objective value
$f$ for difficult multi-modal and deceptive problems.

\subsection{Standard selection schemes (STD)}
The standard selection schemes (abbreviated by STD in the
following), proportionate, truncation, ranking and tournament
selection all favor individuals of higher fitness
\cite{Goldberg:89,Goldberg:91,Blickle:95a,Blickle:97}. This is
also true for less common schemes, like Boltzmann selection
\cite{Maza:93}. The
fitness function is identified with the objective
function (possibly after a monotone transformation).
In linear proportionate selection the probability
of selecting an individual depends linearly on its fitness
\cite{Holland:75}. In truncation selection the $\alpha\%$ fittest
individuals are selected, usually with multiplicity
${1\over\alpha\%}$ to keep the population size fixed
\cite{Muehlenbein:94}.(Linear) ranking selection orders the
individuals according to their fitness. The selection probability
is, then, a (linear) function of the rank \cite{Whitley:89}.
Tournament selection \cite{Baker:85}, which selects the best $l$
out of $k$ individuals has primarily developed for steady-state
EAs, but can be adapted to generation based EAs. All these
selection schemes have the property (and goal!) to increase the
average fitness of a population, i.e.\ to evolve the population
toward higher fitness.

\subsection{The problem of the right selection pressure}
The standard selection schemes STD, together with mutation and
recombination, evolve the population toward higher fitness. If the
selection pressure is too high, the EA gets stuck in a local optimum,
since the genetic diversity rapidly decreases. The suboptimal genetic
material which might help in finding the global optimum is deleted too
rapidly (premature convergence). On the other hand, the selection
pressure cannot be chosen arbitrarily low if we want the EA to be
effective. In difficult optimization problems, suitable population
sizes, mutation and recombination rates, and selection parameters,
which influence the selection intensity, are usually not known
beforehand. Often, constant values are not sufficient at all
\cite{Eiben:99}.  There are various suggestions to dynamically
determine and adapt the parameters
\cite{Eshelman:91,Baeck:91,Herdy:92,Schlierkamp:94}.
Other approaches to preserve genetic diversity are fitness sharing
\cite{Goldberg:87} and crowding \cite{DeJong:75}.
They depend on the proper design of a neighborhood function based on
the specific problem structure and/or coding.  One approach which does
not require a neighborhood function based on the genome is local
mating \cite{Collins:91}, however it has been shown that rapid
takeover can still occur for basic spatial topologies
\cite{Rudolph:00}.  Another approach which has not been widely studied
is preselection \cite{Cavicchio:70}.

We are interested in evolutionary algorithms which do not require
special problem insight (problem specific neighborhood function and/or
coding) and is able to effectively prevent population takeover.  In
this paper we introduce and analyze two potential approaches to this
problem: the Fitness Uniform Selection Scheme (FUSS) and the Fitness
Uniform Deletion Scheme (FUDS).

\subsection{The fitness uniform selection scheme}
FUSS is based on the insight that we are not primarily interested in a
population converging to maximal fitness, but only in a single
individual of maximal fitness.  The scheme automatically creates a
suitable selection pressure and preserves genetic diversity better
than STD. The proposed fitness uniform selection scheme FUSS (see also
Figure \ref{figsel}) is defined as follows: {\em if the lowest/highest
fitness values in the current population $P$ are $f_{min/max}$ we
select a fitness value $f$ uniformly in the interval
$[f_{min},f_{max}]$. Then, the individual $i\in P$ with fitness
nearest to $f$ is selected and a copy is added to $P$, possibly after
mutation and recombination.} We will see that FUSS maintains genetic
diversity better than STD, since a distribution over the fitness
values is used, unlike STD, which all use a distribution over
individuals. Premature convergence is avoided in FUSS by abandoning
convergence at all. Nevertheless there is a selection pressure in FUSS
toward higher fitness.
The probability of selecting a specific individual is proportional
to the distance to its nearest fitness neighbor. In a
population with a high density of unfit and low density of fit
individuals, the fitter ones are effectively favored.

\subsection{The fitness uniform deletion scheme}
We may also preserve diversity through deletion rather than through
selection.  By always deleting from those individuals which have very
commonly occurring fitness values we achieve a population which is
uniformly distributed across fitness values, like with FUSS.  Because
these deleted individuals are ``commonly occurring'' in some sense
this should help preserve population diversity.  Under FUDS the role
of the selection scheme is to govern how actively different parts of
the solution space are searched rather than to move the population as
a whole toward higher fitness.  Thus, like with FUSS, premature
convergence is avoided by abandoning convergence as our goal.  However
as FUDS is only a deletion scheme, the EA still requires a selection
scheme which may require a selection intensity parameter to be set.
Thus we do not necessarily have a parameterless EA, as we do with
FUSS.  Nevertheless due to the impossibility of population collapse
the performance is more robust than usual with respect to variation in
selection intensity.  Thus FUDS is at least a partial solution to the
problem of having to correctly set a selection intensity parameter.

\subsection{Contents}
This paper extends and supersedes the earlier results reported in the
conference papers \cite{Hutter:01fuss}, \cite{Legg:04fussexp} and
\cite{Legg:05fuds}.
Among other things, this paper: extends the previous theoretical
analysis of FUSS and gives the first theoretical analysis of FUDS and
of their performance when combined; presents a new method of analysis
called fitness tree analysis; is the first set of experimental results
which directly compares the two proposed schemes on the same problems
with the same parameters, including when they are used together; gives
the first full analysis of population diversity measurements for FUSS
and in particular extends and corrects some of the earlier speculation
about performance problems in some situations.

The paper is structured as follows:

In {\em Section \ref{secSim}} we discuss the problems of local
optima and population takeover \cite{Goldberg:91} in STD, which
could be lowered by restricting the number of {\em similar}
individuals in a population. As we often do not have an
appropriate functional similarity relation, we define a universal
distance (semi-metric) $d(i,j):=|f(i)-f(j)|$ based on the
available fitness only, which will serve our needs.

Motivated by the universal similarity relation $d$ and by the need to
preserve genetic diversity, we define in {\em Section
\ref{secFuss}} the fitness uniform selection scheme. We
discuss under which circumstances FUSS leads to an (approximate)
fitness uniform population.

Further properties of FUSS are discussed in \emph{Section
\ref{secProp}}, especially, how FUSS creates selection pressure
toward higher fitness and how it preserves diversity better than
STD. Further topics are the equilibrium distribution, the
transformation properties of FUSS under linear and non-linear
transformations of $f$.

Another way to utilize the ability of the universal similarity
relation $d$ to preserve diversity, is to use it to help target
deletion.  This gives us the fitness uniform {\em deletion} scheme
which we define in {\em Section \ref{secFUDS}}.  As this produces a
population which is approximately uniformly distributed across fitness
levels, like with FUSS, many of the properties of FUSS carry over to
an EA using FUDS.  Some of these properties are highlighted in {\em
Section
\ref{secPropFUDS}}.

In {\em Section \ref{secEx}} we theoretically demonstrate, by way of a
simple optimization example, that an EA with FUSS or FUDS can optimize
much faster than with STD. We show that crossover can be effective in
FUSS, even when ineffective in STD. Furthermore, FUSS, FUDS and STD
are compared to random search with and without crossover.

In {\em Section \ref{secTree}} we develop a fitness tree model, which
we believe to cover the essential features of fitness landscapes for
difficult problems with many local optima. Within this model we derive
heuristic expressions for the optimization time of random walk, FUSS,
FUDS and STD.  They are compared, and a worst case slowdown of FUSS
relative to STD is obtained.

There is a possible additional slowdown when including recombination,
as discussed in {\em Section \ref{secCross}}, which can be avoided by
using a scale independent pair selection. It is a ``best'' compromise
between unrestricted recombination and recombination of $d$-similar
individuals only.  It also has other interesting properties when used
without crossover.

To simplify the discussion we have concentrated on the case of
discrete, equi-spaced fitness values. In many practical problems, the
fitness function is continuously valued. FUSS and some of the
discussion of the previous sections is generalized to the continuous
case in {\em Section~\ref{secCont}}.

\emph{Section~\ref{secJfuss}} begins our experimental analysis of
FUSS and FUDS.  In this section we give a detailed account of the EA
software we have used for our experiments, including links to where
the source code can be downloaded.

\emph{Section \ref{secEx2}} examines the empirical performance of
FUSS and FUDS on the artificially constructed deceptive optimization
problem described in Section~\ref{secEx}.  These results confirm the
correctness of our theoretical analysis.

In \emph{Section \ref{secTSP}} we test randomly generated traveling
salesman problems.

In \emph{Section \ref{secSetCover}} we examine the set covering
problem, an NP hard optimization problem which has many real world
applications.

For our final test in \emph{Section \ref{secSAT}} we look at random
CNF3 SAT problems.  These are also NP hard optimization problems.

\emph{Section \ref{secConc}} contains a summary of our results and
possible avenues for future research.

\section{Universal Similarity Relation}\label{secSim}

\subsection{The problem of local optima}
Proportionate, truncation, ranking and tournament are the standard
(STD) selection algorithms used in evolutionary optimization. They
have the following property: if a local optimum $i^{lopt}$ has been
found, the number of individuals with fitness $f^{lopt}=f(i^{lopt})$
tends to increase rapidly. Assume a low mutation and recombination
rate, or, for instance, truncation selection {\em after} mutation and
recombination. Further, assume that it is very difficult to find an
individual fitter than $i^{lopt}$. The population will then degenerate
and will consist mostly of $i^{lopt}$ after a few rounds. This
decreased diversity makes it even less likely that $f^{lopt}$ gets
improved. The suboptimal genetic material which might help in finding
the global optimum has been deleted too rapidly. On the other hand,
too high mutation and recombination rates convert the EA into an
inefficient random search.

\subsection{Possible solution}
Sometimes it is possible to appropriately choose the mutation and
recombination rate and population size by some insight into the nature
of the problem. More often this is a trial and error process, or no
single fixed rate works at all.

A naive fix of the problem is to artificially limit the number of
identical individuals to a significant but small fraction $\eps$.
If the space of individuals $I$ is large, there could be many very
similar (but not identical) individuals of, for instance, fitness
$f^{lopt}$. The EA can still converge to a population containing
only this class of similar individuals, with all others becoming
extinct. In order for the limitation approach to work, one has to
restrict the number of {\em similar} individuals. Significant
contributions in this direction are fitness sharing
\cite{Goldberg:87} and crowding \cite{DeJong:75}.

\subsection{The problem of finding a similarity relation}
If the individuals are coded binary one might use the Hamming distance
as a similarity relation. This distance is consistent with a mutation
operator which flips a few bits. It produces Hamming-similar
individuals, but recombination (like crossover) can produce very
dissimilar individuals w.r.t.\ this measure. In any case, genotypic
similarity relations, like the Hamming distance, depend on the
representation of the individuals as binary strings. Individuals with
very dissimilar genomes might actually be functionally
(phenotypically) very similar. For instance, when most bits are unused
(like introns in genetic programming), they can be randomly disturbed
without affecting the properties of the individual. For specific
problems at hand, it might be possible to find suitable
representation-independent functional similarity relations. On the
other hand, in genetic programming, for instance, it is in general
undecidable whether two individuals are functionally similar.

\subsection{A universal similarity relation}
Here we want to take a different approach. We define the
difference or distance between two individuals as
\beqn
  d(i,j) \;:=\; |f(i)-f(j)|.
\eeqn
The distance is based solely on the fitness function, which is
provided as part of the problem specification.
It is independent of the coding/representation and other problem
details, and of the optimization algorithm (e.g.\ the genetic mutation
and recombination operators), and can trivially be computed from
the fitness values.
If we make the natural assumption that functionally similar
individuals have similar fitness, they are also similar w.r.t.\ the
distance $d$. On the other hand, individuals with very different
coding, and even functionally dissimilar individuals may be
$d$-similar, but we will see that this is acceptable. For instance,
individuals from different local optima of equal height are
$d$-similar.

\subsection{Relation to niching and crowding}
Unlike fitness uniform optimization, diversity control methods like
niching or crowding require a metric $g$ to be defined over the genome
space.  By looking at the relationship between $g$ and $f$ we can
relate these two types of diversity control: We say that a fitness
function $f$ is \emph{smooth} with respect to $g$, if $g( i, j )$
being small implies that $|f(i) - f(j)|$ is also small, that is, $d(
i, j )$ is small.  This implies that if $d( i, j )$ is not small, $g(
i, j )$ also cannot be small.  Thus, if we limit the number of $d$
similar individuals, as we do in fitness uniform optimization, this
will also limit the number of $g$ similar individuals, as is done in
crowding and niching methods.  The advantage of fitness uniform
optimization is that we do not need to know what $g$ is, or to compute
its value.  Indeed, the above argument is true for \emph{any} metric
$g$ on the genome space that $f$ is smooth with respect to.

On the other hand, if the fitness function $f$ is not generally smooth
with respect to $g$, then such a comparison between the methods cannot
be made.  However, in this case an EA is less likely to be effective
as small mutations in genome space with respect to $g$ will produce
unpredictable changes in fitness.

\subsection{Topologies on individual space $I$}
The distance $d:I \times I \to \SetR_0^+$ induced by the fitness
function $f$ is a semi-metric on the individual space $I$ (semi only
because $d(i,j)=0$ for $i\neq j$ is possible). The semi-metric induces
a topology on $I$. Equal fitness suffices to declare two individuals
as $d$-equivalent, i.e. $d$ is a rather small semi-metric in the sense
that the induced topology is rather coarse. We will see that a
non-zero distance between individuals of different fitness is
sufficient to avoiding the population takeover. $d$ induces the
coarsest topology (is the ``smallest'' distance) avoiding population
takeover.

\subsection{The problem of genetic drift}
Besides elitist selection, the other major cause of diversity loss in
a population is genetic drift.  This occurs due to the stochastic
nature of the selection operator breeding some individuals more often
than others.  In a finite population this will cause some individuals
to be replaced which have no close relatives, thus reducing diversity.
Indeed, without a sufficient rate of mutation, eventually a population
will converge on a single genome; even if no selection pressure is
applied.

Although fitness uniform optimization does not attempt to address this
problem, some implications can be drawn.  Clearly, with fitness
uniform optimization a complete collapse in diversity is impossible as
individuals with a wide range of fitness values are always preserved
in the population.  However, within a given fitness level genetic
drift can occur, although the sustained presence of many individuals
in other fitness levels to breed with will reduce this effect.

Theoretical analysis of genetic drift is often performed by
calculating the Markov chain transition matrices to compute the time
for the system to reach an absorption state where all of the
population members have the same genome.  As these results can be
difficult to generalize, an alternative approach has been to measure
genetic drift by measuring the loss in fitness diversity in a
population over time \cite{Rogers:99gd}.  This is interesting as
fitness uniform optimization attempts to maximize the entropy of the
fitness values in the population, producing a very high variance in
population fitness.  Thus, at least according to the second method of
analysis, very little genetic drift would be evident in the
population.

\section{Fitness Uniform Selection Scheme (FUSS)}\label{secFuss}

\subsection{Discrete fitness function}
In this section we propose a new selection scheme, which limits
the fraction of $d$-similar individuals. For simplicity we start
with a fitness function $f:I\to F$ with discrete equi-spaced
values $F=\{f_{min},f_{min}+\eps,f_{min}+2\eps,...,
f_{max}-\eps,f_{max}\}$. We call two individuals $i$ and $j$
$\delta$-similar if $d(i,j)\equiv|f(i)-f(j)|\leq\delta$. The
continuous valued case $F=[f_{min},f_{max}]$ is considered
later. In the following we assume $\delta<\eps$. In this case,
two individuals are $\delta$-similar if and only if they have the
same fitness.

\subsection{The goal}
We have argued that in order to escape local optima, genetic
variety should be preserved somehow. One way is to limit the
number of $\delta$-similar individuals in the population. In an exact
fitness uniform distribution there would be $|P|/|F|$ individuals
for each of the $|F|$ fitness values, i.e.\ each fitness level
would be occupied by a fraction of $1/|F|$ individuals.
The following selection scheme asymptotically transforms any
finite population into a fitness uniform one.

\subsection{The fitness uniform selection scheme (FUSS)}
FUSS is defined as follows: randomly select a fitness value $f$
uniformly from the fitness values $F$.  Then, uniformly at random
select an individual $i\in P$ with fitness $f$. Add another copy of
$i$ to $P$.

\begin{figure}
\centerline{\includegraphics[width=1.0\columnwidth,height=0.6\textheight]{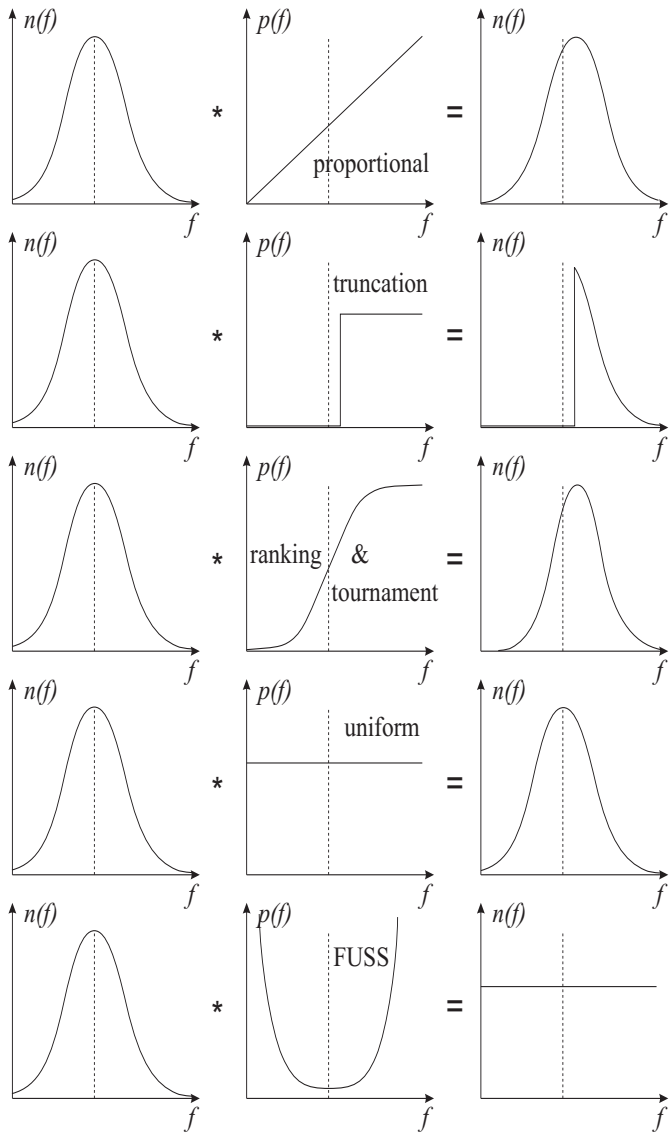}}
\caption{\label{figsel}Effects of proportionate, truncation,
ranking \& tournament, uniform, and fitness uniform (FUSS) selection
on the fitness distribution in a generation based EA.  The left/right
diagrams depict fitness distributions before/after applying the
selection schemes depicted in the middle diagrams.  Note that for
populations with a non-Gaussian distribution of fitness values (left
column), the graph of selection probability vs. fitness for FUSS
(center bottom) can be totally different to that illustrated above,
however the population distribution that results (right bottom) will be
the same.}
\end{figure}

Note the two stage uniform selection process which is very
different from a one step uniform selection of an individual of
$P$ (see Figure \ref{figsel}).
In STD, inertia increases with population size. A large mass of
unfit individuals reduces the probability of selecting fit
individuals. This is not the case for FUSS. Hence, without loss of
performance, we can define a {\em pure model}, in which no
individual is ever deleted; the population size increases with
time. No genetic material is ever discarded and no fine-tuning in
population size is necessary. What may prevent the pure model from
being applied to practical problems are not computation time
issues, but memory problems.
If space becomes a problem we delete random individuals, as is
usually done with a steady state EA.

\subsection{Asymptotically fitness uniform distribution}
The expected number of
individuals per fitness level $f$ after $t$ selections is
$n_t(f)=n_0(f)+t/|F|$, where $n_0(f)$ is the initial distribution.
Hence, asymptotically each fitness level gets occupied uniformly
by a fraction
\beqn
  {n_t(f)\over |P_t|} \;=\;
  {n_0(f)+t/|F|\over |P_0|+t} \;\to\; {1\over |F|}
  \quad\mbox{for}\quad t\to\infty,
\eeqn
where $P_t$ is the population at time $t$. The same limit holds if
each selection is accompanied by uniformly deleting one individual
from the (now constant sized) population.

\subsection{Fitness gaps and continuous fitness}
\begin{figure}
\centerline{\includegraphics[width=1.0\columnwidth,height=0.2\textheight]{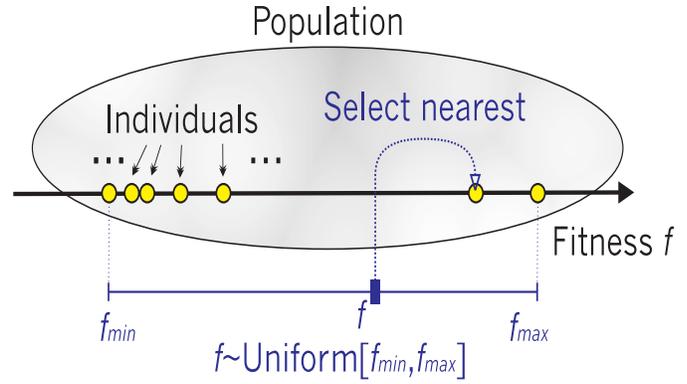}}
\caption{\label{figfuss}If the lowest/highest fitness values
in the current population $P$ are $f_{min/max}$, FUSS selects a
fitness value $f$ uniformly in the interval $[f_{min},f_{max}]$,
then, the individual $i\in P$ with fitness nearest to $f$ is
selected and a copy is added to $P$, possibly after mutation and
recombination.}
\end{figure}

We made two unrealistic assumptions. First, we assumed that each
fitness level is initially occupied. If the smallest/largest
fitness values in $P_t$ are $f_{min/max}^t$ we extend the
definition of FUSS by selecting a fitness value $f$ uniformly in
the interval $[f_{min}^t-\odt\eps,f_{max}^t+\odt\eps]$ and an
individual $i\in P_t$ with fitness nearest to $f$ (see Figure
\ref{figfuss}). This also covers the case when there are missing
intermediate fitness values, and also works for continuous valued
fitness functions ($\eps\to 0$).

\subsection{Mutation and recombination}
The second assumption was that there is no mutation and
recombination. In the presence of a small mutation and/or
recombination rate eventually each fitness level will become
occupied and the occupation fraction is still asymptotically
approximately uniform. For larger rate the distribution will be no
longer uniform, but the important point is that the occupation
fraction of {\em no} fitness level decreases to zero for
$t\to\infty$, unlike for STD.
Furthermore, FUSS selects by construction uniformly in the fitness
levels, even if the levels are not uniformly occupied.

\section{Properties of FUSS}\label{secProp}

\begin{figure}
\centerline{\includegraphics[width=1\columnwidth,height=0.2\textheight]{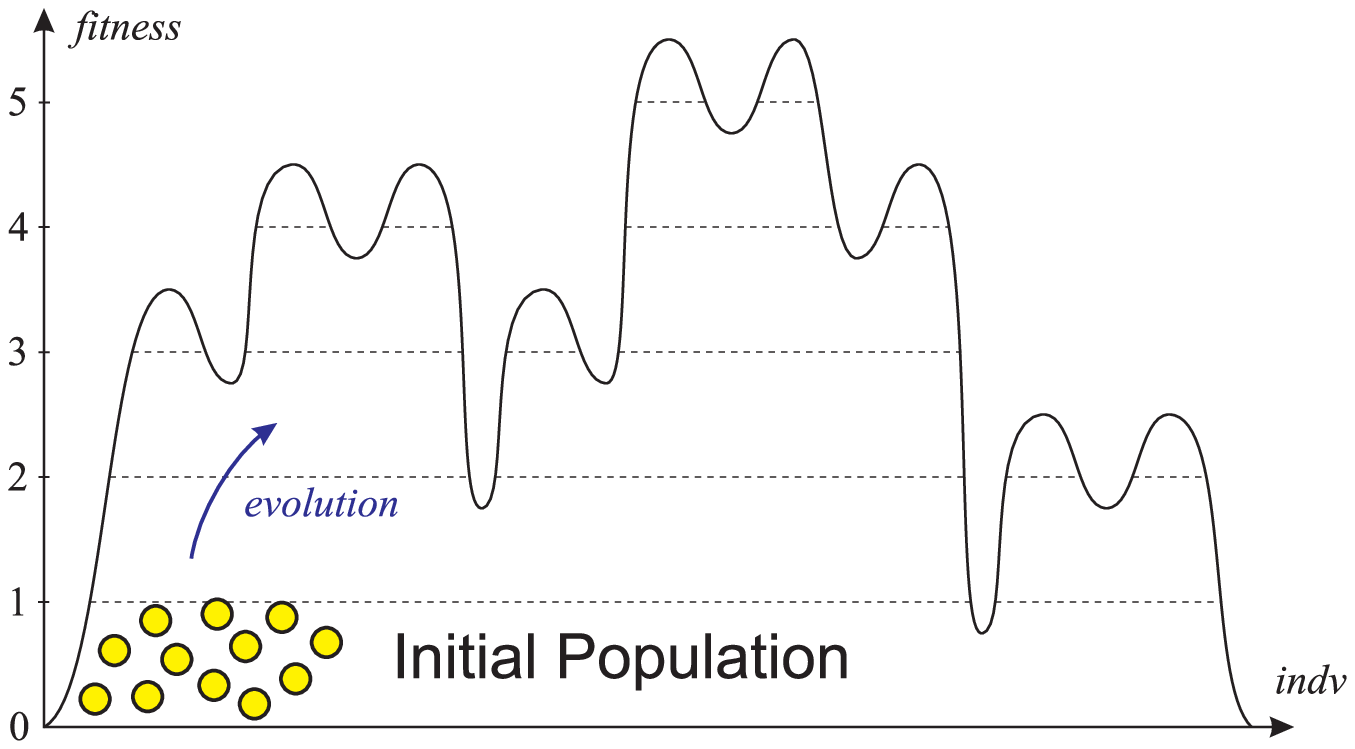}}
\centerline{\includegraphics[width=1\columnwidth,height=0.2\textheight]{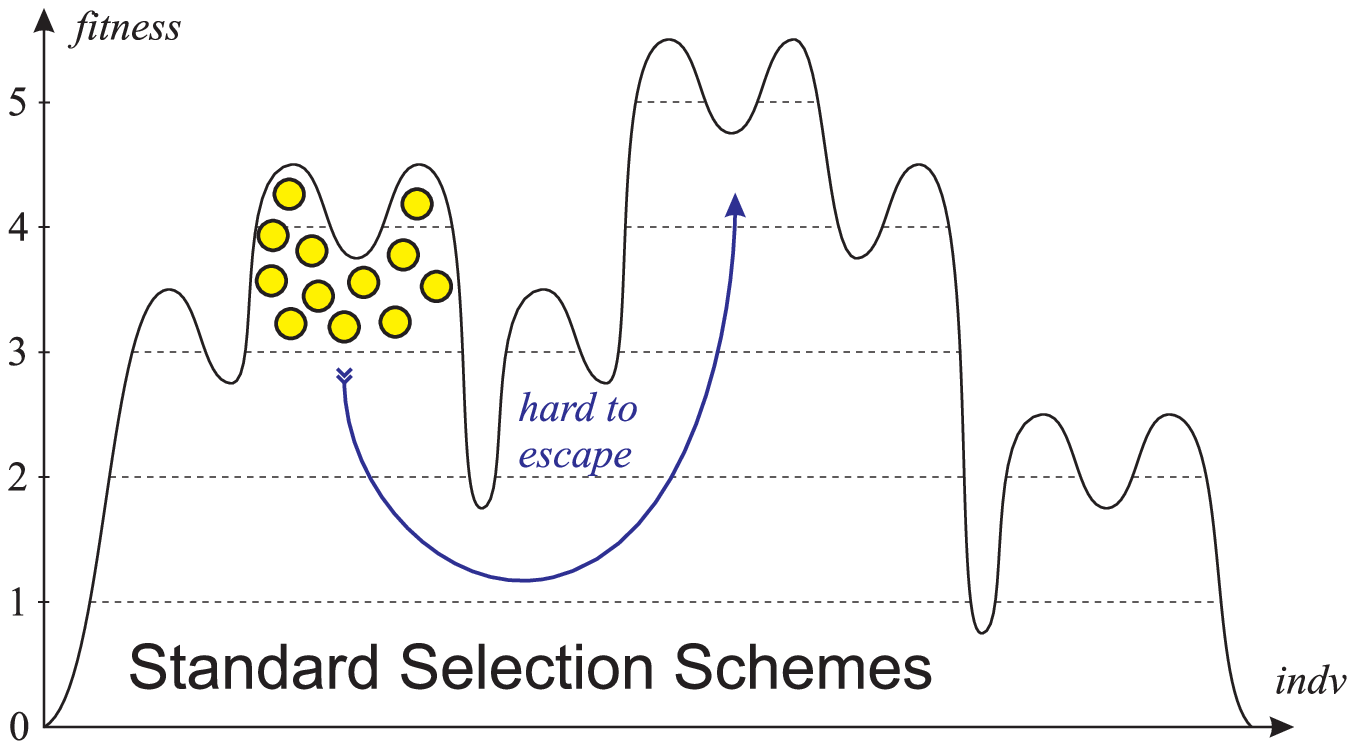}}
\centerline{\includegraphics[width=1\columnwidth,height=0.2\textheight]{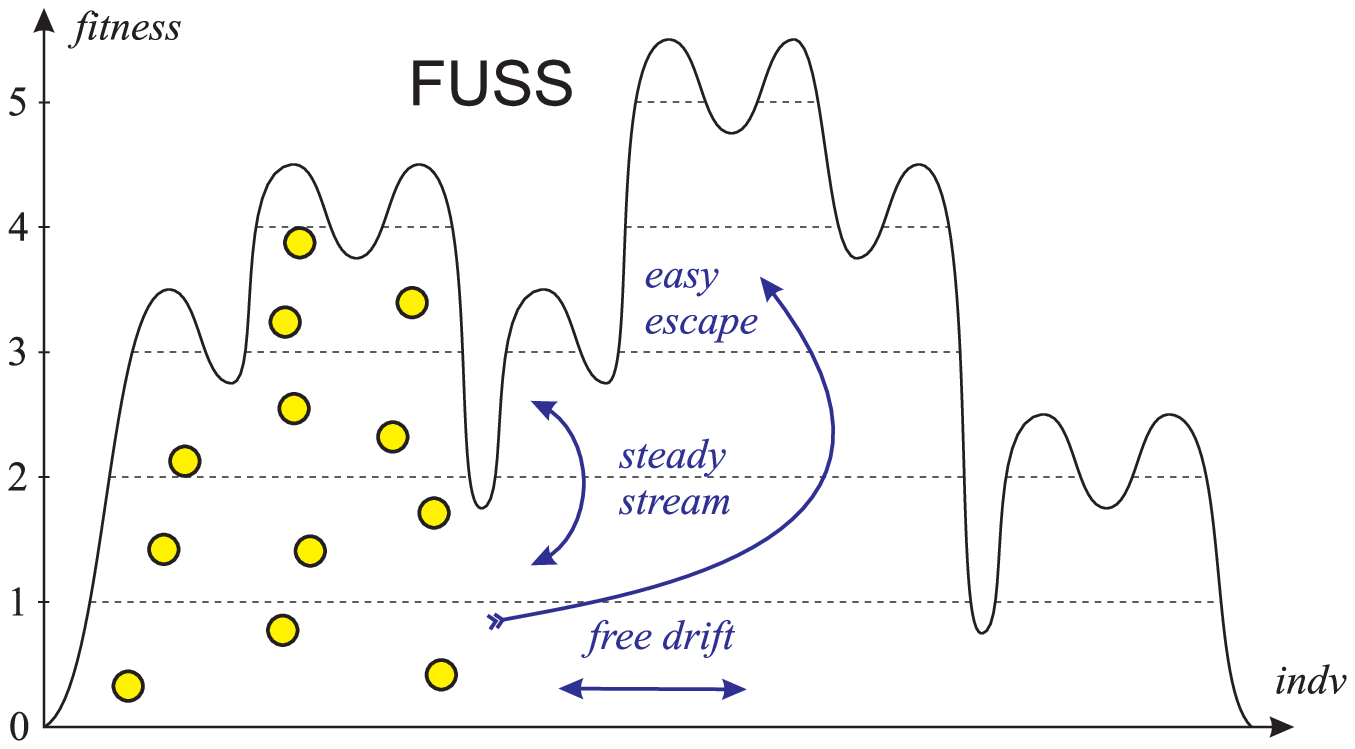}}
\caption{\label{figevolve}Evolution of the population under
FUSS versus standard selection schemes (STD): STD may get stuck
in a local optimum if all unfit individuals were eliminated too
quickly. In FUSS, all fitness levels remain occupied with ``free''
drift within and in-between fitness levels, from which new mutants
are steadily created, occasionally leading to further
evolution in a more promising direction.}
\end{figure}

\subsection{FUSS effectively favors fit individuals}
FUSS preserves diversity better than STD, but the latter have a
(higher) selection pressure toward higher fitness, which is
necessary for optimization. At first glance it seems that there is
no such pressure at all in FUSS, but this is deceiving. As FUSS
selects uniformly in the fitness levels, individuals of low
populated fitness levels are effectively favored. The probability
of selecting a specific individual with fitness $f$ is inversely
proportional to $n_t(f)$ (see Figure \ref{figsel}). In an initial typical
(FUSS) population there are many unfit and only a few fit
individuals. Hence, fit individuals are effectively favored until
the population becomes fitness uniform. Occasionally, a new higher
fitness level is discovered and occupied by a single new
individual, which then, again, is favored.

\subsection{No takeover in FUSS}
With FUSS, takeover of the highest fitness level never happens.  The
concept of takeover time \cite{Goldberg:91} is meaningless for
FUSS. The fraction of fittest individuals in a population is always
small. This implies that the average population fitness is always much
lower than the best fitness. Actually, a large number of fit
individuals is usually not the true optimization goal. A single
fittest individual usually suffices to solve the optimization task.

\subsection{FUSS may also favor unfit individuals}
Note, if it is also difficult to find individuals of low fitness,
i.e.\ if there are only a few individuals of low fitness, FUSS will
also favor these individuals. Half of the time is ``wasted'' in
searching on the wrong end of the fitness scale. This possible
slowdown by a factor of 2 is usually acceptable. In Section
\ref{secEx} we will see that in certain circumstances this
behavior can actually speedup the search. In general, fitness
levels which are difficult to reach, are favored.

\subsection{Distribution within a fitness level}
Within a fitness level there is no selection pressure which could
further exponentially decrease the population in certain regions
of the individual space. This (exponential) reduction is the major
enemy of diversity, which is suppressed by FUSS. Within a fitness
level, the individuals freely drift around (by mutation).
Furthermore, there is a steady stream of individuals into and out
of a level by (d)evolution from (higher)lower levels.
Consequently, FUSS develops an equilibrium distribution which is
nowhere zero. This does not mean that the distribution within a
level is uniform. For instance, if there are two (local) maxima of
same height, a very broad one and a very narrow one, the broad one
may be populated much more than the narrow one, since it is much
easier to ``find''.

\subsection{Steady creation of individuals from every fitness level}
In STD, a wrong step (mutation) at some point in evolution might cause
further evolution in the wrong direction. Once a local optimum has
been found and all unfit individuals were eliminated it is very
difficult to undo the wrong step. In FUSS, all fitness levels remain
occupied from which new mutants are steadily created, occasionally
leading to further evolution in a more promising direction (see Figure
\ref{figevolve}).

\subsection{Transformation properties of FUSS}
FUSS (with continuous fitness) is independent of a scaling and a
shift of the fitness function, i.e.\ FUSS($\tilde f$) with $\tilde
f(i):=a\cdot f(i)+b$ is identical to FUSS($f$). This is true
even for $a<0$, since FUSS searches for maxima {\em and}
minima, as we have seen. It is not independent of a non-linear
(monotone) transformation unlike tournament, ranking and
truncation selection. The non-linear transformation properties are
more like the ones of proportionate selection.

\section{Fitness Uniform Deletion Scheme (FUDS)}\label{secFUDS}

For a steady state evolutionary algorithm each cycle of the system
consists of both selecting which individual or individuals to
crossover and mutate, and then selecting which individual is to be
deleted in order to make space for the new child.  The usual deletion
scheme used is \emph{random deletion} as this is neutral in the sense
that it does not bias the distribution of the population in any way
and does not require additional work to be done, such as evaluating
the similarity of individuals based on their genes.  Another common
strategy is to use an elitist deletion scheme.

Here we propose to use the similarity semi-metric $d$ defined in
Section \ref{secSim} to achieve a uniform distribution across fitness
levels, like with FUSS, except that we achieve this by selectively
deleting those members of the population which have very commonly
occurring fitness values.  Of course this leaves the selection scheme
unspecified, indeed we may use any standard selection scheme such as
tournament selection in combination with FUDS.  It also means that we
lose one of the nice features of FUSS as we now need to manually tune
the selection intensity for our application --- FUSS of course is
parameterless.  Nevertheless it allows us to give many FUSS like
properties to an existing EA using a standard selection scheme with
only a minor modification to the deletion scheme.

The intuition behind why FUDS preserves population diversity is very
simple: If an individual has a fitness value which is very rare in the
population then this individual almost certainly contains unique
information which, if it were to be deleted, would decrease the total
population diversity.  Conversely, if we delete an individual with
very commonly occurring fitness then we are unlikely to be losing
significant diversity.  Presumably most of these individuals are
common in some sense and likely exist in parts of the solution space
which are easy to reach.  Thus the fitness uniform deletion strategy
is now clear: Only delete individuals with very commonly occurring
fitness values as these individuals are less likely to contain
important genetic diversity.

Practically FUDS is implemented as follows.  Let $f_{min}$ and
$f_{max}$ be the minimum and maximum fitness values possible for a
problem, or at least reasonable upper and lower bounds.  We divide the
interval $[f_{min}, f_{max}]$ into a collection of subintervals of
equal length $\{ [f_{min}, f_{min} + a ), [f_{min} + a, f_{min} + 2a
), \ldots, [f_{max}-a, f_{max}] \}$ which we call \emph{fitness
levels}.  As individuals are added to the population their fitness is
computed and they are placed in the set of individuals corresponding
to the fitness level they belong to.  Thus the number of individuals
in each fitness level describes how common fitness values within this
interval are in the current population.  When a deletion is required
the algorithm locates the fitness level with the greatest number of
individuals and then deletes a random individual from this level.  In
the case where multiple fitness levels have maximal size the lowest of
these levels is used.

If the number of fitness levels is chosen too low, say 5 levels, then
the resulting model of the distribution of individuals across the
fitness range will be too coarse.  Alternatively if a large number of
fitness levels is used with a very small population the individuals
may become too thinly spread across the fitness levels.  While in
these extreme cases this could affect the performance of FUDS, in
practice we have found that the system is not very sensitive to the
setting of this parameter.  If $n$ is the population size then setting
the number of fitness levels to be $\sqrt{n}$ is a good rule of thumb.

For discrete valued fitness functions there is a natural lower bound on
the interval length $a$ because below a certain value there will be
more intervals than unique fitness values.  Of course this cannot
happen when the fitness function is continuous.  Other than this small
technical detail, the two cases are treated identically.

As FUDS spreads the individuals out across a wide range of fitness
values, for small populations the EA may become inefficient as only a
few individuals will have relatively high fitness.  For problems which
are not deceptive this is especially true as there will be little
value in having individuals in the population with low to medium
fitness.  Of course these are not the kinds of problems for which FUDS
was designed.  In practice we have always used populations of between
250 and 5,000 individuals and have not observed a decline in
performance relative to random deletion at the lower end of this
range.

An alternative implementation that avoids discretization is to choose
the two individuals that have the most similar fitness and delete one
of them. An efficient implementation keeps a list of the individuals
ordered by their fitness along with an ordered list of the distances
between the individuals.  Then in each cycle one of the two
individuals with closest fitness to each other is selected for
deletion.  Although the performance of this algorithm was better than
random deletion, it was not as good as the implementation of FUDS
using bins.  We conjecture that the reason for this is as follows:
When there are just a few very fit individuals in the population it is
quite likely that they will be highly related to each other and have
very similar fitness.  This means that if we delete the individuals
with most similar fitness it is likely that many of the very fit
individuals will be deleted.  However with the bins approach this will
not happen as there are typically few individuals in the high fitness
bins.  Thus, although deleting one of the closest individuals in terms
of fitness might preserve diversity well, it also changes the pressure
on the population distribution over fitness levels.  This small change
in distribution dynamics appears to reduce performance in practice.

\section{Properties of FUDS}\label{secPropFUDS}

As FUDS uniformly distributes the population across fitness levels,
like FUSS does, many of the key properties of FUSS also carry over to
an EA that is using a standard selection scheme (STD) combined with
FUDS deletion.

\subsection{No takeover in FUDS}
Under FUDS the takeover of the highest fitness level, or indeed any
fitness level, is impossible.  This is easy to see because as soon as
any fitness level starts to dominate, all of the deletions become
focused on this level until it is no longer the most populated fitness
level.  As a by-product, this also means that individuals on
relatively unpopulated fitness levels are preserved.

\subsection{Steady creation of individuals from every fitness level}
Another similarity with FUSS is the steady creation of individuals on
many different fitness levels.  This occurs because under FUDS some
individuals on each fitness level are always kept.  This makes it
relatively easy for the EA to find its way out of local optima as it
keeps on exploring evolutionary paths which do not at first appear to
be promising.

\subsection{Robust performance with respect to selection intensity}
Because FUDS is only a deletion scheme, we still need to choose a
selection scheme for the EA.  Of course this selection scheme may then
require us to set a selection intensity parameter.  While this is not
as desirable as FUSS, which has no such parameter, at least with FUDS
we expect the performance of the system to be less sensitive to the
correct setting of this parameter.  For example, if the selection
intensity is set too high the normal problem is that the population
rushes into a local optimum too soon and becomes stuck before it has
had a chance to properly explore the genotype space for other
promising regions.  However, as we noted above, with FUDS a total
collapse in population diversity is impossible.  Thus much higher
levels of selection intensity may be used without the risk of
premature convergence.

In some situations if very low section intensity is used along with
random deletion, the population tends not to explore the higher areas
of the fitness landscape at all.  This can be illustrated by a simple
example.  Consider a population which contains 1,000 individuals.
Under random deletion all of these individuals, including the highly
fit ones, will have a 1 in 1,000 chance of being deleted in each cycle
and so the expected life time of an individual is 1,000 deletion
cycles.  Thus if a highly fit individual is to contribute a child of
the same fitness or higher, it must do so reasonably quickly.  However
for some optimization problems the probability of a fit individual
having such a child when it is selected is very low, so low in fact
that it is more likely to be deleted before this happens.  As a result
the population becomes stuck, unable to find individuals of greater
fitness before the fittest individuals are killed off.

The usual solution to this problem is to increase the selection
intensity because then the fit individuals are selected more often and
thus are more likely to contribute a child of similar or greater
fitness before they are deleted.  Another is to change the deletion
scheme so that these individuals live longer.  This is what happens
with FUDS as rare fit individuals are not deleted.  Effectively it
means that with FUDS we can often use much lower selection intensity
without the population becoming stuck.

\subsection{Transformation properties of FUDS}
While with FUDS we have the added complication of having to choose the
number of subintervals with which to break up the fitness values, this
number is only a function of the population size and distributional
characteristics of the problem.  Thus any linear transformation of the
fitness function has no effect on FUDS.  However, non-linear
transformations will affect performance.

\subsection{Problem and representation independence}
Because FUDS only requires the fitness of individuals, the method is
completely independent of the problem and genotype representation,
i.e.\ how the individuals are coded.

\subsection{Simple implementation and low computational cost}
As the algorithm is simple and the fitness function is given as part
of the problem specification, FUDS is very easy to implement and
requires few computational resources.

\section{A Simple Example}\label{secEx}

In the following, we use a simple example problem to compare the
performance of fitness uniform selection (FUSS), random search (RAND)
and standard selection (STD), each used both with and without
recombination.  We also examine the performance of standard selection
when used with the fitness uniform deletion scheme (FUDS).  We regard
this problem as a prototype for deceptive multi-modal functions.  The
example demonstrates how FUSS and FUDS can be superior to RAND and STD
in some situations.  More generic situations will be considered in
Section \ref{secTree}.  An experimental analysis of this problem
appears in Section~\ref{secEx2}.

\subsection{Simple 2D example}
Consider individuals $(x,y)\in I:=[0,1]\times[0,1]$,
which are tuples of real numbers, each coordinate in the interval $[0,1]$.
The example models individuals possessing up to 2 ``features''.
Individual $i$ possesses feature $I_1$ if
$i\in I_1:=[a,a+\Delta]\times[0,1]$, and feature $I_2$
if $i\in I_2:=[0,1]\times[b,b+\Delta]$.
The fitness function $f:I\to\{1,2,3\}$ is defined as
\beqn
  f(x,y) = \left\{
  \begin{array}{l}
    1 \quad\mbox{if}\quad (x,y)\in I_1\backslash I_2, \\
    2 \quad\mbox{if}\quad (x,y)\in I_2\backslash I_1, \\
    3 \quad\mbox{if}\quad (x,y)\not\in I_1\cup I_2, \\
    4 \quad\mbox{if}\quad (x,y)\in I_1\cap I_2. \\
  \end{array}\right.
\parbox{2cm}{\hfill \unitlength=0.6mm
\begin{picture}(45,45)
\scriptsize
\put(5,5){\vector(0,1){40}}
\put(5,5){\vector(1,0){40}}
\put(20,5){\line(0,1){35}}
\put(25,5){\line(0,1){35}}
\put(40,5){\line(0,1){35}}
\put(5,15){\line(1,0){35}}
\put(5,20){\line(1,0){35}}
\put(5,40){\line(1,0){35}}
\put(22.5,17.5){\makebox(0,0)[cc]{4}}
\put(12.5,30){\makebox(0,0)[cc]{3}}
\put(22.5,30){\makebox(0,0)[cc]{1}}
\put(32.5,30){\makebox(0,0)[cc]{3}}
\put(32.5,10){\makebox(0,0)[cc]{3}}
\put(12.5,10){\makebox(0,0)[cc]{3}}
\put(12.5,17.5){\makebox(0,0)[cc]{2}}
\put(32.5,17.5){\makebox(0,0)[cc]{2}}
\put(22.5,10){\makebox(0,0)[cc]{1}}
\put(44,2.5){\makebox(0,0)[cc]{$x$}}
\put(22.5,2.5){\makebox(0,0)[cc]{$\Delta$}}
\put(20,3.5){\makebox(0,0)[cc]{$a$}}
\put(2.5,17.5){\makebox(0,0)[cc]{$\Delta$}}
\put(4,14.5){\makebox(0,0)[cc]{$b$}}
\put(40,3){\makebox(0,0)[cc]{1}}
\put(3.5,40){\makebox(0,0)[cc]{1}}
\put(2.5,44){\makebox(0,0)[cc]{$y$}}
\put(22.5,42.5){\makebox(0,0)[cc]{$f(x,y)$}}
\end{picture}
}
\eeqn
We assume $\Delta\ll 1$. Individuals with neither of the two features
($i\in I\backslash(I_1\cup I_2)$) have fitness $f=3$. These ``local
$f=3$ optima'' occupy most of the individual space $I$, namely a
fraction $(1-\Delta)^2$. It is disadvantageous for an individual to
possess only one of the two features ($i\in(I_1\backslash I_2)\cup
(I_2\backslash I_1)$), since $f=1$ or 2 in this case. In combination
($i\in I_1\cap I_2)$), the two features lead to the highest fitness,
but the global maximum $f=4$ occupies the smallest fraction $\Delta^2$
of the individual space $I$. With a fraction $\Delta(1-\Delta)$, the
$f=1/f=2$ minima are in between. The example has sort of an XOR
structure, which is hard for many optimizers.

\subsection{Random search}
Individuals are created uniformly in the unit square. The ``local
optimum'' $f=3$ is easy to ``find'', since it occupies nearly the
whole space. The global optimum $f=4$ is difficult to find, since it
occupies only $\Delta^2\ll 1$ of the space. The expected time, i.e.\
the expected number of individuals created and tested until one with
$f=4$ is found, is $T_{RAND}={1\over\Delta^2}$. Here and in the
following, the ``time'' $T$ is defined as the expected number of
created individuals until the {\it first} optimal individual (with
$f=4$) is found. $T$ is neither a takeover time nor the number of
generations (we consider steady-state EAs).

\subsection{Random search with crossover}
Let us occasionally perform a recombination of individuals in the
current population. We combine the $x$-coordinate of one uniformly
selected individual $i_1$ with the $y$ coordinate of another
individual $i_2$. This crossover operation maintains a uniform
distribution of individuals in $[0,1]^2$. It leads to the global
optimum if $i_1\in I_1$ and $i_2\in I_2$. The probability of
selecting an individual in $I_i$ is
$\Delta(1-\Delta)\approx\Delta$ (we assumed that the global
optimum has not yet been found). Hence, the probability that $I_1$
crosses with $I_2$ is $\Delta^2$. The time to find the global
optimum by random search including crossover is still
$\sim{1\over\Delta^2}$ ($\sim$ denotes asymptotic proportionality).

\subsection{Mutation}
The result remains valid (to leading order in ${1\over\Delta}$)
if, instead of a random search, we uniformly select an individual
and mutate it according to some probabilistic, sufficiently mixing
rule, which preserves uniformity in $[0,1]$. One popular such
mutation operator is to use a sufficiently long binary
representation of each coordinate, like in genetic algorithms, and
flip a single bit. For simplicity we assume in the following a
mutation operator which replaces with probability $\odt/\odt$ the
first/second coordinate by a new uniform random number. Other
mutation operators which mutate with probability $\odt/\odt$ the
first/second coordinate, preserve uniformity, are sufficiently
mixing, and leave the other coordinate unchanged (like the
single-bit-flip operator) lead to the same scaling of $T$ with
$\Delta$ (but with different proportionality constants).

\subsection{Standard selection with crossover}
The $f=1$ and $f=2$ individuals contain useful building
blocks, which could speedup the search by a suitable selection and
crossover scheme. Unfortunately, the standard selection schemes
favor individuals of higher fitness and will diminish the
$f=1/f=2$ population fraction. The probability of
selecting $f=1/f=2$ individuals is even smaller than in
random search. Hence $T_{STD}\sim{1\over\Delta^2}$. Standard
selection does not improve performance, even not in combination
with crossover, although crossover is well suited to produce the
needed recombination.

\subsection{FUSS}
At the beginning, only the $f=3$ level is occupied and
individuals are uniformly selected and mutated. The expected time
until an $f=1$ or $f=2$ individual in $I_1\cup I_2$ is created is
$T_1\approx{1\over \Delta}$ (not ${1\over 2\Delta}$, since only
one coordinate is mutated). From this time on FUSS will select one
half(!) of the time the $f=1/f=2$ individual(s) and only the
remaining half the abundant $f=3$ individuals. When level
$f=1$ {\em and} level $f=2$ are occupied, the selection
probability is ${1\over 3}+{1\over 3}$ for these levels.
With probability $\odt$ the
mutation operator will mutate the $y$ coordinate of an individual
in $I_1$ or the $x$ coordinate of an individual in $I_2$ and
produces a new $f=1/2/4$ individual. The relative probability
of creating an $f=4$ individual is $\Delta$. The expected time
to find this global optimum from the $f=1/f=2$ individuals, hence,
is $T_2=[({1\over 2}...{2\over 3})\times{1\over
2}\times\Delta]^{-1}$. The total expected time is
$T_{FUSS}\approx T_1+T_2= {4\over\Delta}...{5\over\Delta}\ll
{1\over\Delta^2}\sim T_{STD}$. FUSS is much faster by exploiting
unfit $f=1/f=2$ individuals. This is an example where (local)
minima can help the search. Examples where a low local maxima
can help in finding the global maximum, but where standard
selection sweeps over too quickly to higher but useless local
maxima, can also be constructed.

\subsection{FUSS with crossover}
The expected time until an $f=1$ individual in $I_1$ and an
$f=2$ individual in $I_2$ is found is $T_1\sim{1\over
\Delta}$, even with crossover. The probability of selecting an
$f=1/f=2$ individual is ${1\over 3}/{1\over 3}$. Thus, the
probability that a crossing operation crosses $I_1$ with $I_2$ is
$({1\over 3})^2$. The expected time to find the global optimum
from the $f=1/f=2$ individuals, hence, is $T_2=9\cdot O(1)$,
where the $O(1)$ factor depends on the frequency of crossover
operations. This is far faster than by STD, even if the
$f=1/f=2$ levels were local maxima, since to get a high standard
selection probability, the level has first to be taken over, which
itself needs some time depending on the population size. In FUSS a
single $f=1$ and a single $f=2$ individual suffice to
guarantee a high selection probability and an effective crossover.
Crossover does not significantly decrease the {\em total} time
$T_{FUSSX}\approx T_1+T_2\sim {1\over \Delta}+O(9)$, but for a
suitable 3D generalization we get a large speedup by a factor of
${1\over\Delta}$.

\subsection{FUDS with crossover}
Assume that initially all of the individuals have $f=3$ and that we
are using random selection.  For any mutation the probability of the
child being in $I_1 \cup I_2$ is $\Delta$.  Until $I_1 \cup I_2$
becomes quite full FUDS will never delete individuals from these
areas.  Furthermore if an individual in $I_1 \cup I_2$ is mutated then
the mutant will also be in $I_1 \cup I_2$ with probability
$\frac{1}{2}( 1 + \Delta) \gg \Delta$.  Therefore while most of the
population has $f=3$ we can lower bound the probability of a new child
being in $I_1 \cup I_2$ by $\Delta$.  It then follows that if $P$ is
the size of the population we can upper bound the expected time for
$I_1 \cup I_2$ to contain half the total population by $\frac{P}{2}
\frac{1}{\Delta}
\propto \frac{1}{\Delta}$.  Once this occurs (and most likely well
before this point) crossover will produce an individual with $f=4$
almost immediately by crossing a member of $I_1$ with a member of
$I_2$.  Thus $T_{FUDS} \propto \frac{1}{\Delta} \ll \frac{1}{\Delta^2}
\sim T_{STD}$.  This gives FUDS when used with random selection scaling
characteristics which are similar to FUSS.  If we use a selection
scheme with higher intensity our bound on the expected time for half
the population to have $f=3$ remains unchanged as the bound holds in
the worst case situation where only individuals with $f=3$ are
selected.  However higher selection intensity makes the final
crossover required to find an individual with $f=4$ less likely.  For
moderate levels of selection intensity this is clearly not a
significant factor and more importantly it is O(1) and independent of
$\Delta$.  Thus the order of scaling for $T_{FUDS}$ is just
$\frac{1}{\Delta}$ for this difficult problem, which is the same as
$T_{FUSSX}$.

\subsection{Simple 3D example}
We generalize the 2D example to D-dimensional individuals
$\vec x\in[0,1]^D$ and a
fitness function
\beqn
  f(\vec x) \;:=\; (D+1)\!\cdot\!\prod_{d=1}^D\chi_d(\vec x)\;
  - \max_{1\leq d\leq D} d\!\cdot\!\chi_d(\vec x)\; +D+1,
\eeqn
where $\chi_d(\vec x)$ is the characteristic function of feature
$I_d$
\beqn
  \chi_d(\vec x) \;:=\; \left\{
  \begin{array}{l}
    1 \quad\mbox{if}\quad a_i\leq x_i\leq a_i+\Delta, \\
    0 \quad\mbox{else.} \\
  \end{array}\right.
\eeqn
For $D=2$, $f$ coincides with the 2D example. For $D=3$,
the fractions of $[0,1]^3$ where $f=1/2/3/4/5$ are approximately
$\Delta^2/\Delta^2/\Delta^2/1/\Delta^3$.
With the same line of reasoning we get the following expected
search times for the global optimum:
\beqn
  T_{RAND}\sim T_{STD}\sim {1\over\Delta^3},
\eeqn \beqn
  T_{FUSS}\sim {1\over\Delta^2},\quad
  T_{FUSSX}\sim T_{FUDS} \sim {1\over\Delta}.
\eeqn
This demonstrates the existence of problems where FUSS is much faster
than RAND and STD, and where crossover can give a further boost to
FUSS, even when it is ineffective in combination with STD.

\section{Fitness-Tree Analysis}\label{secTree}

\begin{figure}
\centerline{\includegraphics[width=\columnwidth,height=0.25\textheight]{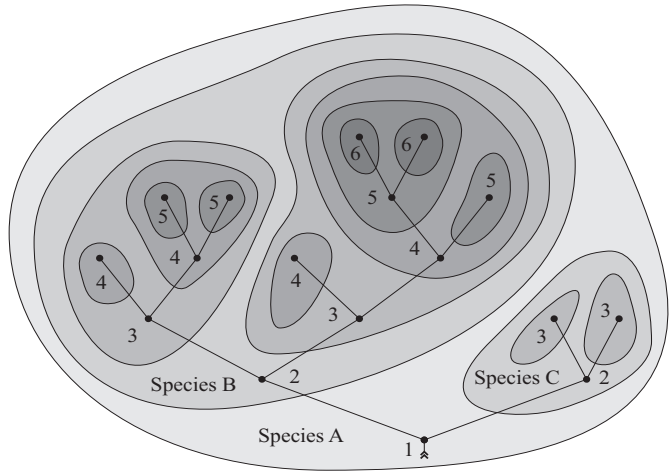}}
\caption{\label{figtree}Generic 2D fitness landscape with
evolution tree. Each connected slice represents a species. A species
is also symbolized by a node in the slice. The number in a slice and
near a node is the fitness value of the species. If individuals from
one species can evolve to individuals of another species, the nodes are
connected by a solid line. Altogether, they form the fitness tree. The
branching factor $b$ is $2$ and the number of species per fitness
level $s$ is $4$ for intermediate fitness values (3,4,5).}
\end{figure}

subsection{The fitness tree model}
A general, problem independent comparison of the various
optimization algorithms is difficult. We are interested in the
performance for difficult fitness landscapes with many local
optima.

\begin{figure}
\centerline{\includegraphics[width=\columnwidth,height=0.25\textheight]{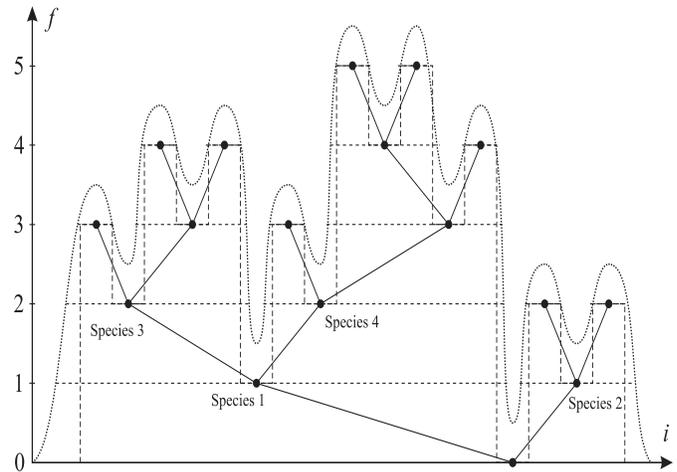}}
\caption{\label{figtree1d}
Generic fitness function with evolution tree. Individuals which
are evolutionary neighbors are connected by a dashed line. They
belong to the species indicated by a node on the dashed
line. A species which can evolve from another is connected to
it by a solid line. The smooth curve
visualizes (somewhat misleading, since the fitness is discrete)
the fitness function with many local maxima. \vspace{1.1em}}
\end{figure}

We only consider mutation; recombination is discussed in the next
section. The evolutionary neighborhood (not to be confused with
$d$-similarity) of an individual $i$ is defined as the set of
individuals that can be created from $i$ by a single
mutation\footnote{We have ``small'' mutations in mind, e.g.\ single
bit flips, not macro mutations, which connect {\em all}
individuals.}. Two individuals $i$ and $j$ with the same fitness are
defined to belong to the same {\em species} if there is a finite
sequence of mutations which transforms $i$ into $j$ {\em and} all
individuals of the sequence also have fitness $f(i)=f(j)$. Each
fitness level is partitioned in this way into disjoint species. We say
a species of fitness $f+\eps$ can {\em evolve} from a species of
fitness $f$, if there is a mutation which transforms an individual
from the latter species to one of the former. Those species are
connected by an edge in Figures \ref{figtree} and
\ref{figtree1d}. A species is said to be {\em promising} if it
{\em can} evolve to the global optimum $f_{max}$.

\subsection{Additional definitions and simplifying assumptions}
\begin{itemize}\parskip=0ex\parsep=0ex\itemsep=0ex
\item[i)] Evolution which skips fitness levels is ignored, and also
devolution to species of lower fitness other than the
primordial species.
\item[ii)] Random individuals have
lowest fitness $f_{min}$ with high probability, and there is
only one species of fitness $f_{min}$.
\item[iii)] There is a fixed branching factor $b$, i.e.\ each species
can evolve into $b$ improved species, or represents a local optimum
from which no further evolution is possible.
\item[iv)] There is a single global optimum
$f_{max}$ (or $b$ optima to be consistent with the previous item).
\item[v)] There are $s$ different species per fitness level (except
near $f_{min}$ and $f_{max}$ where there must be fewer to be consistent
with the previous items).
\item[vi)] The probability $p$ that an individual evolves to a higher
fitness is very small. In most cases a mutation keeps an
individual within its species or devolves it.
\item[vii)] The probability to evolve to one of the offspring species is
uniform, i.e.\ $1/b$ for all offspring species.
\end{itemize}

We have the feeling that this picture covers the essential
features of fitness landscapes for difficult problems. The
qualitative conclusions we will draw should still hold when some
or all of the additional simplifying assumptions are violated.

\subsection{Example}
Consider the case of individuals, which are real-valued $D$
dimensional vectors, i.e.\ $I=\SetR^D$. Let the fitness function
$\tilde f$ be continuous and positive with many local maxima,
which tends to zero for large arguments. This covers a large range
of physical optimization problems. Mutation shall be local in
$\SetR^D$, i.e. $||i_{original}-i_{mutated}||\ll D$. As FUSS and
the fitness tree model is only defined for discrete fitness
functions, we discretize $\tilde f$ to
$f:=\,_\lfloor{1\over\tilde\eps}\tilde f_\rfloor$, which is
acceptable for sufficiently small $\tilde\eps$. A typical fitness
landscape for $D=2$ and $D=1$ together with their fitness tree are
depicted in Figures \ref{figtree} and \ref{figtree1d}. Since
mutation is a local operation, each species is a (possibly
multiply punched) connected slice ($D$-dimensional sub-volume) and
evolution can only occur from $f$ to $f+1$ ($\eps=1$). Assumption
(i) is generally satisfied. The special fitness landscapes
depicted in Figures \ref{figtree} and \ref{figtree1d} also satisfy
(ii,iii,iv,v) with $b=2$ and $s=4$.

\subsection{Random walk}
Consider a mutation induced random walk of a single individual.  Due
to the low evolution probability $p\ll 1$, most of the time will be
spent on individuals of the lowest fitness $f_{min}$. As evolution is
a tree, there is only one evolution sequence which leads to the global
optimum. At each evolution step, the correct offspring species (out of
$b$) has to be evolved. The probability of an evolution step in the
right direction, hence, is $p/b$. $|F|$ evolution steps are necessary
to reach $f_{max}$. Therefore, the expected time to find the global
maximum by random walk is $T_{RW}\approx(b/p)^{|F|}$. Random walk is
very slow; it is exponential in the number of fitness levels $|F|$ to
a very large basis $b/p$.

\subsection{FUSS}
Assume that $L$ fitness levels from $f_{min}$ to $f$ are occupied.
The probability that FUSS selects an individual of fitness $f$ is
$1/L$. Under this additional assumption that the occupation of species
within one fitness level is approximately uniform most of the time,
the probability of selecting an individual of the promising species,
which can evolve to the global optimum, is $1/s$. The probability of
an evolution step in the right direction is $p/b$ as in the random
walk case. Hence, the total expected time for an evolution in the
right direction is $L\cdot s\cdot b/p$. The total time
$T_{FUSS}\approx\odt|F|^2\cdots\cdot b/p$ for an evolution from $L=1$
to the global optimum $L=|F|$ is obtained by summation over
$L=1...|F|$.

\subsection{FUDS}
A similar analysis can be applied to FUDS.  Assume again that the $L$
fitness levels from $f_{min}$ to $f$ are occupied and that the
occupation of species within each fitness level is approximately
uniform most of the time.  Because FUDS tends to spread the population
out, like FUSS, this assumption is not unreasonable.  As FUDS is only
a deletion scheme we must also specify a selection scheme.  For our
analysis we will take a very simple elitist selection scheme that half
of the time selects an individual from the highest fitness level, and
the other half of the time selects an individual from a lower level.
It follows then that the probability of selecting a promising species
is $1/2s$ and the probability that this then results in an
evolutionary step in the right direction is $p/b$.  Thus the total
expected time for an evolutionary step in the right direction is $2
\cdot s \cdot b/p$.  Therefore by summation the total expected time to
evolve to the global optimum is $T_{FUDS} \approx 2|F| \cdot s \cdot
b/p$.  Of course this analysis rests on our choice of selection scheme
and the assumptions about the uniformity of the population that we
have made.  When FUDS is used with selection schemes which are very
greedy these uniformity assumptions will likely be violated and less
favorable bounds could result.

\subsection{Standard selection}
We assumed a fixed number of $s$ species per fitness level and $0$ or
$b$ offspring species. This implies that only a fraction of $1/b$
species can evolve to higher fitness. We assume that fitness level
$f$ has been taken over, i.e.\ most individuals have fitness $f$. The
probability of evolution is $p$. A significant fraction (for
simplicity we assume most) of the $|P|$ individuals must evolve to the
next fitness level before evolution with a relevant rate can occur to
the next to next level. Hence, the time to take over the next fitness
level is roughly $|P|\cdot b/p$.  As there are $|F|$ fitness levels,
the total time is $T_{STD}
\approxgeq |F|\cdot|P|\cdot b/p$.

We wrote $\approxgeq$ as we have made two significant favorable
assumptions. In order to ensure convergence, the promising species in
the current fitness level has to be occupied. If we assume a uniform
occupation of species within one fitness level, as for FUSS, this
means that all species of the current fitness level have to be
populated. As there are $s$ species, $|P|$ has to be at least $s$,
which can be quite large. On the other hand, STD linearly slows down
with $|P|$, unlike FUSS. Hence, there is a trade-off in the choice of
$|P|$.

More serious is the following problem. Assume that the first
individual evolved with fitness $f+\eps$ is one in a non-promising
species $a$. Due to selection pressure it might happen that
species $a$ takes over the whole population before all (or at
least the promising) species with fitness $f+\eps$ can evolve from
the ones of fitness $f$. The probability to find the global
optimum in the worst case scenario, where at each level only one
species is occupied, is $(1/b)^{|F|}$. This is the original
problem of the loss of genetic diversity discussed at the outset,
which lead to the invention of FUSS.

Every other fix the authors are aware of only seems to diminish the
problem, but does not solve it. One fix is to repeatedly restart
the EA, but the huge number of $b^{|F|}$ restarts might be
necessary. The time is exponential in $|F|$ like for random walk
but with a smaller basis $b$. The true time is expected to be
somewhere in between $|F|\cdot|P|\cdot b/p$ and this worst
case analysis, although an unfavorable setting may never reach the
global optimum ($T_{STD}=\infty$ in this case).

\subsection{Performance comparison}
The times $T_{FUSS}$, $T_{FUDS}$ and $T_{STD}$ should be regarded, at
best, as rules of thumb, since the derivation was rather heuristic due
to the list of assumptions. The quotient is more reliable:
\beqn
  {T_{FUSS}\over T_{STD}} \quad\approxleq\quad
  {|F|\!\cdot\!s\over 2|P|} \quad\approxleq\quad
  \odt|F| \quad\leq\quad |F|,
\eeqn
and
\beqn
  {T_{FUDS}\over T_{STD}} \quad\approx\quad
  {\frac{s }{|P|}} \quad\approx\quad 1.
\eeqn

We will give a more direct argument in Section \ref{secCross} that
the slowdown of FUSS relative to STD is at most $|F|$.

Finally, a truism has been recovered, namely that an EA can, under
certain circumstances, be much faster than random walk, that is,
$T_{RW}\gg T_{FUSS}, T_{FUDS}, T_{STD}$.

\section{Scale-Independent Selection and Recombination}\label{secCross}

\subsection{Worst case analysis}
We now want to estimate the maximal possible slowdown of FUSS
compared to STD.
Let us assume that all individuals in STD have fitness $f$, and
once one individual with fitness $f+\eps$ has been found,
takeover of level $f+\eps$ is quick. Let us assume that this
quick takeover is actually good (e.g.\ if there are no local maxima).
The selection probability of individuals of same fitness is equal.
For FUSS we assume individuals in the range of $f_{min}$ and $f$.
Uniformity is {\em not} necessary. In the worst case, a selection of
an individual of fitness $<f$ never leads to an individual of
fitness $\geq f$, i.e.\ is always useless. The probability of selecting
an individual with fitness $f$ is $\geq{1\over|F|}$.
At least every $|F|th$ FUSS selection corresponds to a STD
selection. Hence, we expect a maximal slowdown by a factor of
$|F|$, since FUSS ``simulates'' STD statistically every $|F|th$
selection.
It is possible to construct problems where this slowdown occurs
(unimodal function, local mutation $x\to x\pm\eps$, no
crossover). Gradient ascent would be the algorithm of choice in this
case. On the other hand, we have not observed this slowdown in our
simple 2D example and the TSP experiments, where FUSS outperformed STD
in solution quality/time (see the experimental results in
Section~\ref{secEx2}). Since real world problems often lie in between
these extreme cases it is desirable to modify FUSS to cope with simple
problems as well, without destroying its advantages for complex
objective functions.

\subsection{Quadratic slowdown due to recombination}
We have seen that $T_{FUSS}\leq|F|\cdot T_{STD}$. In the
presence of recombination, a {\em pair} of individuals has to be
selected. The probability that FUSS selects {\em two} individuals
with fitness $f$ is $\geq{1\over|F|^2}$. Hence, in the worst case,
there could be a slowdown by a factor of $|F|^2$ --- for {\em
independent} selection we expect
$T_{FUSS}\leq|F|^2\cdot T_{STD}$. This potential quadratic
slowdown can be avoided by selecting one fitness value at random,
and then two individuals of this single fitness value. For this
{\em dependent} selection, we expect
$T_{FUSS}\leq|F|\cdot T_{STD}$. On the other hand,
crossing two individuals of different fitness can also be
advantageous, like the crossing of $f=1$ with $f=2$
individuals in the 2D example of Section \ref{secEx}.

\subsection{Scale independent selection}
A near optimal compromise is possible: a high selection
probability $p(f)\sim 1$ if $f\approx f_{max}$ and $p(f)\sim
{1\over|F|}$ otherwise. A ``scale
independent'' probability distribution $p(f)\sim{1\over|f_{max}-f|}$
is appropriate for this.
We define
\beq\label{ptscale}
  p(f) \;:=\; {c\over\ln|F|}\cdot
  {1\over {1\over\eps}|f_{max}-f|+1}.
\eeq
The $+1$ in the denominator has been added to regularize the
expression for $f=f_{max}$. The factor $c/\ln|F|$ ensures
correct normalization ($\sum_f p(f)=1$). By using $\ln{b+1\over
a}\leq\sum_{i=a}^b{1\over i}\leq\ln{b\over a-1}$, one can show
that
$
  {\ln|F|\over 1+\ln|F|} \leq c \leq 1
$
i.e.\ $c\to 1$ for $|F|\to\infty$. In the following we assume
$|F|\geq 3$, i.e.\ $c\geq \odt$.
Apart from a minor additional logarithmic suppression of order
$\ln|F|$ we have the desired behavior $p(f)\sim 1$
for $f\approx f_{max}$ and $p(f)\sim {1\over|F|}$ otherwise:
\beqn
  p(f_{max}-m\eps) \geq {1\over 2\ln|F|} \cdot
  {1\over m+1},
\eeqn
\beqn
  p(f) \geq {1\over 2\ln|F|} \cdot
  {1\over |F|} \quad\forall\,f
\eeqn
During optimization, the minimal/maximal fitness of an individual in
population $P_t$ is $f_{min/max}^t$. In the definition of $p$ one has
to use $F_t:=\{f_{min}^t,f_{min}^t+\eps,...,f_{max}^t\}$ instead of
$F$, i.e.\ $|F|$ replaced with
$|F_t|={1\over\eps}(f_{max}^t-f_{min}^t)+1\leq|F|$. So (\ref{ptscale})
can not be achieved by a static re-parametrization of fitness $f$
replaced with $g(f)$. Furthermore the important idea of sampling from
a fitness level instead of individuals directly is still
maintained. The only difference now is that the population will no
longer converge to a fitness uniform one but to one with distribution
$p(f)$ which is biased toward higher fitness but still never converges
to a fittest individual. In the worst case, we expect a small slowdown
of the order of $\ln|F|$ as compared to FUSS, as well as compared to
STD.

\subsection{Scale independent pair selection}
It is possible to (nearly) have the best of independent and
dependent selection: a high selection probability $p(f,f')\sim
{1\over|F|}$ if $f\approx f'$ and $p(f,f')\sim {1\over|F|^2}$
otherwise, with uniform marginal $p(f)={1\over|F|}$. The idea
is to use a strongly correlated joint distribution for selecting a
fitness pair. A ``scale independent'' probability distribution
$p(f,f')\sim{1\over|f-f'|}$ is appropriate. We define the joint
probability $\tilde p(f,f')$ of selecting two individuals of
fitness $f$ and $f'$ and the marginal $\tilde p(f)$ as
\begin{equation} \label{ptjoint}
  \tilde p(f,f') \;:=\; {1\over 2|F|\ln|F|}\cdot
  {1\over {1\over\eps}|f\!-\!f'|+1},
\end{equation}
\[
  \tilde p(f) \;:=\; \sum_{f'\in F}\tilde p(f,f')
  = \sum_{f'\in F}\tilde p(f',f).
\]

We assume $|F|\geq 3$ in the following. The $+1$ in the
denominator has been added to regularize the expression for
$f=f'$. The factor $(2|F|\ln|F|)^{-1}$ ensures correct
normalization for $|F|\to\infty$. More precisely, using
$\ln{b+1\over a}\leq\sum_{i=a}^b{1\over i}\leq\ln{b\over a-1}$,
one can show that
\beqn
  1-{\textstyle{1\over\ln|F|}} \;\leq\;
  \sum_{f,f'\in F}\tilde p(f,f') \;\leq\; 1,\quad
  \odt \;\leq\; |F|\!\cdot\!\tilde p(f) \;\leq\; 1,
\eeqn
i.e.\ $\tilde p$ is not strictly normalized to $1$ and the
marginal $\tilde p(f)$ is only approximately (within a factor of 2)
uniform. The first defect can be corrected by appropriately
increasing the diagonal probabilities $\tilde p(f,f)$. This also
solves the second problem.
\beq\label{pjoint}
  p(f,f') \;:=\; \left\{
  \begin{array}{ll}
    \tilde p(f,f') & \mbox{for}\quad f\neq f' \\
    \tilde p(f,f')+[{1\over|F|}-\tilde p(f)] &
    \mbox{for}\quad f=f' \
  \end{array}
\right.
\eeq

\subsection{Properties of $p(f,f')$}
$p$ is normalized to $1$ with uniform marginal
\[
  p(f):= \sum_{f'\in F} p(f,f') = {1\over|F|},
\]
\[
  \sum_{f,f'\in F} p(f,f') =
  \sum_{f\in F} p(f) = 1,
\]
\[
  p(f,f')\geq \tilde p(f,f').
\]
Apart from a minor additional logarithmic suppression of order
$\ln|F|$ we have the desired behavior $p(f,f')\sim {1\over|F|}$
for $f\approx f'$ and $p(f,f')\sim {1\over|F|^2}$ otherwise:
\[
  p(f,f\pm m\eps) \geq {1\over 2\ln|F|} \cdot
  {1\over m+1} \cdot {1\over|F|},
\]
\[
  p(f,f') \geq {1\over 2\ln|F|} \cdot
  {1\over |F|^2}.
\]
During optimization, the minimal/maximal fitness of an individual in
population $P_t$ is $f_{min/max}^t$. In the definition of $p$ one has
to use $F_t:=\{f_{min}^t,f_{min}^t+\eps,...,f_{max}^t\}$ instead of
$F$, i.e.\ $|F|$ replaced with
$L:=|F_t|={1\over\eps}(f_{max}^t-f_{min}^t)+1\leq|F|$.

\subsection{Scale-Independent Deletion}
Just as the selection scheme FUSS has its dual in the deletion scheme
FUDS, we can likewise create the dual of Scale-Independent Selection
in the form of Scale-Independent Deletion.  Thus rather than targeting
deletion from the population so that the distribution becomes flat, as
we do with FUDS, we now define a convex curve $g$ which is peaked at
the fittest individual in the population and delete the population
down so that it follows the shape of this curve.  This retains some of
the advantages of FUDS, for example the population cannot collapse to
just a few fitness levels, and yet it recognizes that for many
problems it is useful to bias the population distribution toward fit
individuals.  Of course such problems are less deceptive than the kind
that FUSS and FUDS are intended for.

\section{Continuous Fitness Functions}\label{secCont}

\subsection{Effective discretization scale}
Up to now we have considered a discrete valued fitness function
with values in $F=\{f_{min},f_{min}+\eps,...,f_{max}\}$.
In many practical problems, the fitness function is continuous
valued with $F=[f_{min},f_{max}]$. We generalize FUSS, and some of
the discussion of the previous sections to the continuous case by
replacing the discretization scale $\eps$ by an effective
(time-dependent) discretization scale $\hat\eps$. By construction, FUSS shifts
the population toward a more uniform one. Although the fitness
values are no longer equi-spaced, they still form a discrete set
for finite population $P$. For a fitness uniform distribution, the
average distance between (fitness) neighboring individuals is
${1\over|P_t|-1}(f^t_{max}-f^t_{min})=:\hat\eps$. We
define $\hat
F_t:=\{f^t_{min},f^t_{min}+\hat\eps,...,f^t_{max}\}$.
$|\hat F_t| = {1\over\hat\eps}(f^t_{max}-f^t_{min})+1 =
|P_t|$.

\subsection{FUSS}
Fitness uniform selection for a continuous valued function has already
been mentioned in Section \ref{secFuss}. We just take a uniform random
fitness $f$ in the interval
$[f_{min}^t-\odt\hat\eps,f_{max}^t+\odt\hat\eps]$.
Independent and dependent fitness pair selection as described in
the last section works analogously. An $\hat\eps=0$ version of
correlated selection does not exist; a non-zero $\hat\eps$ is
important. A discrete pair $(f,f')$ is drawn with probability
$p(f,f')$ as defined in (\ref{ptjoint}) and (\ref{pjoint}) with
$\eps$ and $F$ replaced by $\hat\eps$ and $\hat F_t$. The
additional suppression $\ln|\hat F|=\ln|P_t|$ is small for all
practically realizable population sizes.
In all cases an individual with fitness nearest to $f$ ($f'$) is
selected from the population $P$ (randomly if there is
more than one nearest individual).

If we assume a fitness uniform distribution then our worst case bound
of $T_{FUSS}\approxleq\sum_{t=1}^{T_{STD}}|P_t|$ is plausible, since
the probability of selecting the best individual is approximately
$|P_t|$. For constant population size we get a bound
$T_{FUSS}\approxleq|P|\cdot T_{STD}$. For the preferred non-deletion
case with population size $|P_t|=t$ the bound gets much worse
$T_{FUSS}\approxleq\odt T_{STD}^2$.
This possible (but not necessary!) slowdown has similarities to
the slowdown problems of proportionate selection in later
optimization stages.
The species definition in Section \ref{secTree} has to be relaxed
by allowing mutation sequences of individuals with
$\hat\eps$-similar fitness.
Larger choices of $\hat\eps$ may be favorable if the standard
choice causes problems.

\subsection{FUDS}
Fitness uniform deletion already requires the range of the fitness
function to be broken up into a finite number of intervals.  While for
discrete valued fitness functions the intervals may correspond to the
unique values of the fitness function, this is not a requirement.
Indeed if the population is small and the fitness function has a large
number of possible values then a more coarse discretization is
necessary.  Continuous valued fitness functions can therefore be
treated in exactly the same way and do not cause any special problems.
In fact they are slightly simpler in that we are now free to choose
the discretization as fine as we like without being limited by the
number of possible fitness values.  Of course, like in the discrete
case, we still must choose a discretization which is appropriate given
the size of the population.

\section{The EA Test System}\label{secJfuss}

To test FUSS and FUDS we have implemented an EA test system in Java.
The complete source code along with the test problems presented in
this paper and basic usage instructions can be downloaded from
\cite{Legg:website}.  The EA model we have chosen for our tests is the
so called ``steady state'' model as opposed to the more usual
``generational'' model.  In a generational EA in each generation we
select an entirely new population based on the old population.  The
old population is then simply discarded.  Under the steady state model
that we use, each step of the optimization adds and removes just one
individual at a time.  Specifically the process occurs as follows:
Firstly an individual is selected by the
\emph{selection scheme} and then with a certain probability another
individual is also selected and the \emph{crossover operator} is
applied to produce a new individual.  Then with another probability a
\emph{mutation operator} is applied to produce the child individual
which is then added to the population.  We refer to the probability of
crossing as the \emph{crossover probability} and the probability of
mutating following a crossover as the \emph{mutation probability}.  In
the case where no crossover takes place the individual is always
mutated to ensure that we are not simply adding a clone of an existing
individual into the population.  Finally, an individual must be
deleted in order to keep the population size constant.  This
individual is selected by the \emph{deletion scheme}.  The deletion
scheme is important as it has the power to bias the population in a
similar way to the selection scheme.

Our task in this paper is to experimentally analyze how FUSS performs
relative to other selection schemes and how FUDS performs relative to
other deletion schemes.  Because any particular run of a steady state
EA requires both a selection and a deletion scheme to be used, there
are many possible combinations that we could test.  We have narrowed
this range of possibilities down to just a few that are commonly used.

Among the selection schemes, tournament selection is one of the
simplest and most commonly used and we consider it to be roughly
representative of other standard selection schemes which favor the
fitter individuals in the population; indeed in the case of tournament
size 2 it can be shown that tournament selection is equivalent to the
linear ranking selection scheme \cite[Sec.2.2.4]{Hutter:92cfs}.  With
tournament selection we randomly pick a group of individuals and then
select the fittest individual from this group.  The size of the group
is called the \emph{tournament size} and it is clear that the larger
this group is the more likely we are to select a highly fit individual
from the population.  At some point in the future we may implement
other standard selection schemes to broaden our comparison, however we
expect the performance of these schemes to be at best comparable to
tournament selection when used with a correctly tuned selection
intensity.

Among the deletion schemes one of the most commonly used in steady
state EAs is random deletion.  The rational for this is that it is
neutral in the sense that it does not skew the distribution of the
population in any way.  Thus whether the population tends toward high
or low fitness etc.\ is solely a function of the selection scheme and
its settings.  Of course random deletion, unlike FUDS, makes no effort
to preserve diversity in the population as all individuals have an
equal chance of being removed.  In this paper we will compare FUDS
against random deletion as this is the standard deletion schemes in
situations where it is difficult or impossible to directly measure the
similarity of individuals based on their genomes.

\begin{figure*}[t]
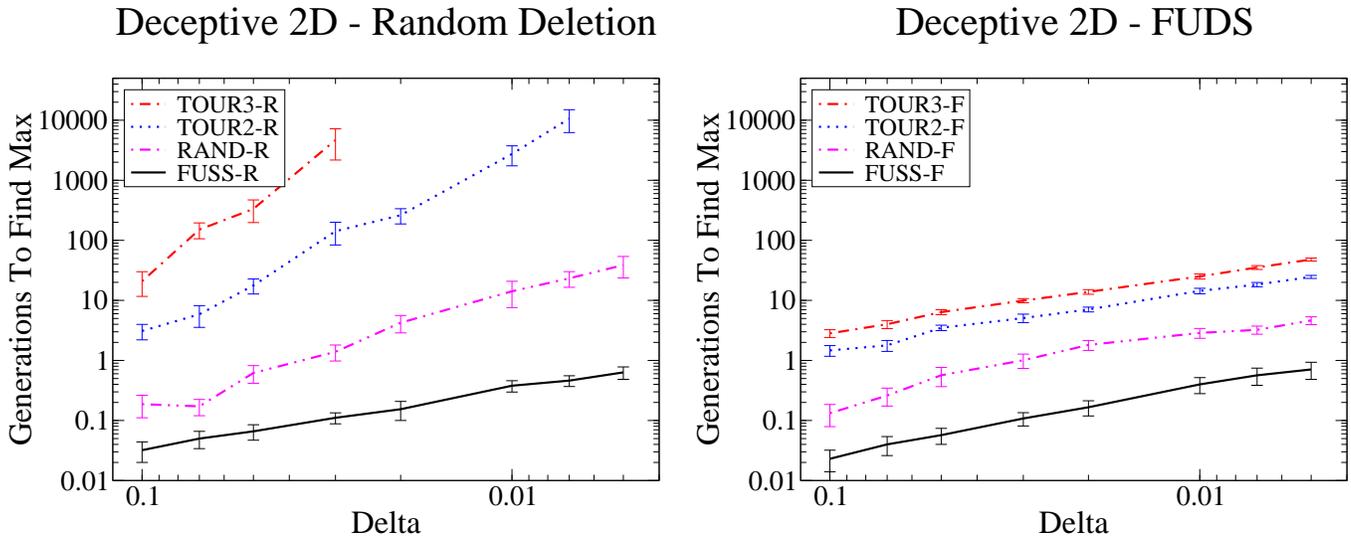

\includegraphics[width=0.485\textwidth]{Deceptive-R.eps}
\includegraphics[width=0.01\textwidth]{space.eps}
\includegraphics[width=0.485\textwidth]{Deceptive-F.eps}
\caption{\label{SimpleProb} With random deletion (left graph) FUSS
significantly outperforms TOURx and RAND.  By switching to FUDS (right
graph) the performance of TOURx and RAND now scale the same as FUSS.}
\end{figure*}

When reporting test results we will adopt the following notation:
TOUR2 means tournament selection with a tournament size of 2.
Similarly for TOUR3, TOUR4 and so on.  Under random selection, denoted
RAND, all members of the population have an equal probability of being
selected.  This is sometimes called uniform selection.  When a graph
shows the performance of tournament selection over a range of
tournament sizes we will simply write TOURx.  Naturally FUSS indicates
the fitness uniform selection scheme.  To indicate the deletion scheme
used we will add either the suffix \mbox{-R} or \mbox{-F} to indicate
random deletion or FUDS respectively.  Thus, TOUR10-R is tournament
selection with a tournament size of 10 used with random deletion,
while FUSS-F is FUSS selection used with FUDS deletion.

The important free parameters to set for each test are the population
size, and the crossover and mutation probabilities.  Good values for
the crossover and mutation probabilities depend on the problem and
must be manually tuned based on experience as there are few
theoretical guidelines on how to do this.  For some problems
performance can be quite sensitive to these values while for others
they are less important.  Our default values are 0.5 for both as this
has often provided us with reasonable performance in the past.

For each test we ran the system multiple times with the same mutation
and crossover probabilities and the same population size.  The only
difference was which selection and deletion schemes were used by the
code.  Thus even if our various parameters, mutation operators etc.\
were not optimal for a given problem, the comparison is still fair.
Indeed we often deliberately set the optimization parameters to
non-optimal values in order to compare the robustness of the systems.

As a steady state optimizer operates on just one individual at a time,
the number of cycles within a given run can be high, perhaps 100,000
or more.  In order to make our results more comparable to a
generational optimizer we divide this number by the size of the
population to give the approximate number of generations.
Unfortunately the theoretical understanding of the relationship
between steady state and generational optimizers is not strong.  It
has been shown that under the assumption of no crossover the effective
selection intensity using tournament selection with size 2 is
approximately twice as strong under a steady state EA as it is with a
generational EA \cite{Rogers:99}.  As far as we are aware a similar
comparison for systems with crossover has not been performed.

Depending on the purpose of a test run, different stopping criteria
were applied.  For example, in situations where we wanted to graph how
rapidly different strategies converged with respect to generations, it
made sense to fix the number of generations.  In other situations we
wanted to stop a run once the optimizer appeared to have become stuck,
that is, when the maximum fitness had not improved after some
specified number of generations.  In any case we explain for each test
the stopping criterion that has been used.

In order to generate reliable statistics we ran each test multiple
times; typically 30 times but sometimes up to 100 times if the results
were noisy.  From these runs we then calculated the mean performance
as well as the sample standard deviation and from this the standard
error in our estimate of the mean.  This value was then used to
generate the 95\% confidence intervals which appear as error bars on
the graphs.

\section{A Deceptive 2D Problem}\label{secEx2}

The first problem we examine is the simple but highly deceptive 2D
optimization problem which was theoretically analyzed in
Section~\ref{secEx}.  As in the theoretical analysis, we set up the
mutation operator to randomly replace either the $x$ or $y$ position
of an individual and the crossover to take the $x$ position from one
individual and the $y$ position from another to produce an offspring.
The size of the domain for which the function is maximized is just
$\delta^2$ which is very small for small values of $\delta$, while the
local maxima at fitness level 3 covers most of the space.  Clearly the
only way to reach the global maximum is by leaving this local maximum
and exploring the space of individuals with lower fitness values of 1
or 2.  Thus, with respect to the mutation and crossover operators we
have defined, this is a deceptive optimization problem as these
partitions mislead the EA \cite{Forrest:93}.

For this test we set the maximum population size to 1,000 and made 20
runs for each $\delta$ value.  With a steady state EA it is usual to
start with a full population of random individuals.  However for this
particular problem we reduced the initial population size down to just
10 in order to avoid the effect of doing a large random search when we
created the initial population and thereby distorting the scaling.
Usually this might create difficulties due to the poor genetic
diversity in the initial population.  However due to the fact that any
individual can mutate to any other in just two steps this is not a
problem in this situation.  Initial tests indicated that reducing the
crossover probability from 0.5 to 0.25 improved the performance
slightly and so we have used the latter value.

The first set of results for the selection schemes used with random
deletion appear in the left graph of Figure~\ref{SimpleProb}.  As
expected, higher selection intensity is a significant disadvantage for
this problem.  Indeed even with just a tournament size of 3 the number
of generations required to find the maximum became infeasible to
compute for smaller values of $\delta$.  Our results confirm the
theoretical scaling orders of $1\over\delta^2$ for TOUR2-R, and
$1\over\delta$ for FUSS-R, as predicted in Section~\ref{secEx}.  Be
aware that this is a log-log scaled graph and so the different slopes
indicate significantly different orders of scaling.

In the second set of tests we switch from random deletion to FUDS.
These results appear in the right graph of Figure~\ref{SimpleProb}.
We see that with FUDS as the deletion scheme the scaling improves
dramatically for RAND, TOUR2 and TOUR3.  Indeed they are now of the
same order $\frac{1}{\delta}$ as FUSS, as predicted in
Section~\ref{secEx}.  This shows that for very deceptive problems much
higher levels of selection intensity can be applied when using FUDS
rather than random deletion.  The performance of FUSS-R is very
similar to that of FUSS-F.  This is not surprising as the population
distribution under FUSS already tends to be approximately uniform
across fitness levels and thus we expect the effect of FUDS to be
quite weak.

Although this problem was artificially constructed, the results
clearly demonstrate how FUSS and FUDS can dramatically improve
performance in some situations.

\section{Traveling Salesman Problem}\label{secTSP}

\begin{figure*}[t]
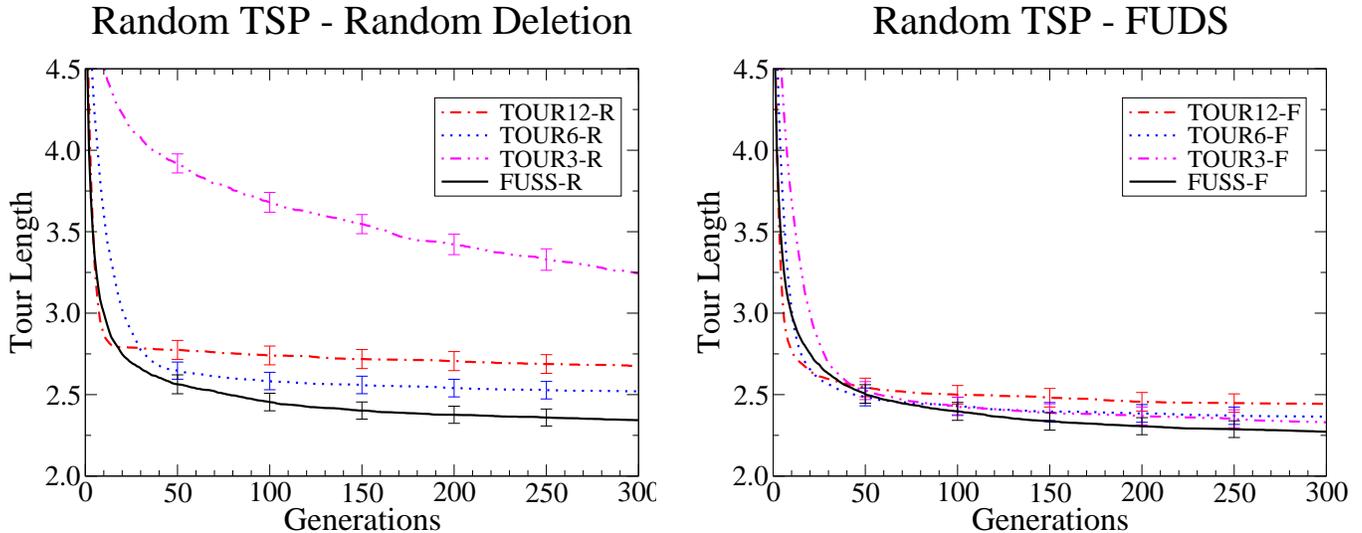

\includegraphics[width=0.485\textwidth]{DTSPI-300gen-R.eps}
\includegraphics[width=0.01\textwidth]{space.eps}
\includegraphics[width=0.485\textwidth]{DTSPI-300gen-F.eps}
\caption{\label{DTSP-1} TOUR3-R converged too slowly while TOUR12-R
converged prematurely and became stuck.  TOUR6-R appears to be about
the correct tournament size for this problem, however it is still
inferior to FUSS-R.  With FUDS all of the selection schemes performed
well though FUSS was still the best.}
\end{figure*}

A well known optimization problem is the so called Traveling Salesman
Problem (TSP).  The task is to find the shortest Hamiltonian cycle
(path) in a graph of $N$ vertexes (cities) connected by edges of
certain lengths.  There exist highly specialized population based
optimizers which use advanced mutation and crossover operators and are
capable of finding paths less than one percent longer than the optimal
path for up to $10^7$ cities
\cite{Lin:73,Martin:96,Johnson:97,Applegate:00}.  As our goal is only
to study the relative performance of selection and deletion schemes,
having a highly refined implementation is not important.  Thus the
mutation and crossover operators we used were quite simple: Mutation
was achieved by just switching the position of two of the cities in
the solution, while for crossover we used the partial mapped crossover
technique \cite{Goldberg:85}.  Fitness was computed by taking the
reciprocal of the tour length.

For our first set of tests we used randomly generated TSP problems,
that is, the distance between any two cities was chosen uniformly from
the unit interval $[0,1]$.  We chose this as it is known to be a
particularly deceptive form of the TSP problem as the usual triangle
inequality relation does not hold in general.  For example, the
distance between cities $A$ and $B$ might be $0.1$, between cities $B$
and $C$ $0.2$, and yet the distance between $A$ and $C$ might be
$0.8$.  The problem still has some structure though as efficient
partial solutions tend to be useful building blocks for efficient
complete tours.

For this test we used random distance TSP problems with 20 cities and
a population size of 1000.  We found that changing the crossover and
mutation probabilities did not improve performance and so these have
been left at their default values of 0.5.  Our stopping criterion was
simply to let the EA run for 300 generations as this appeared to be
adequate for all of the methods to converge and allowed us to easily
graph performance versus generations.

The first graph in Figure~\ref{DTSP-1} shows each of the selection
schemes used with random deletion.  We see that TOUR3-R has
insufficient selection intensity for adequate convergence while
TOUR12-R quickly converges to a local optimum and then becomes stuck.
TOUR6-R has about the correct level of selection intensity for this
problem and population size.  FUSS-R however initially converges as
rapidly as TOUR12-R but avoids becoming stuck in local optima.  This
suggests improved population diversity.  The performance curve for
FUSS-R is impressive, especially considering that it is parameterless.

At first it might seem surprising that the maximum fitness with FUSS
climbs very quickly for the first 20 generations, especially
considering that FUSS makes no attempt to increase the average fitness
in the population.  However we can explain this very rapid rise in
solution fitness by considering a simple example.  Consider a
situation where there is a large number of individuals in a small band
of fitness levels, say 1,000 with fitness values ranging from 50 to
70.  Add to this population one individual with a fitness value of 73.
Thus the total fitness range contains 24 values.  Whenever FUSS picks
a random point from 72 to 73 inclusive this single individual with
maximal fitness will be selected.  That is, the probability that the
single fittest individual will be selected is 2/24 = 0.083.  In
comparison under TOUR12 the probability that the fittest individual is
selected is the same as the probability that it is picked for the
sample of 12 elements used for the tournament, which is approximately,
12/1000 = 0.012.  Thus the probability of the fittest individual in
the population being selected is higher under FUSS than under TOUR12
and so the maximum fitness would rise quickly to start with.

Previously in \cite{Legg:04fussexp} we speculated that this may have
been responsible for performance problems that we had observed with
FUSS in some situations.  However further experimentation has shown
that very rapid rises in maximal fitness are quite rare and are also
very shortly lived when they do occur --- too short to cause any
significant diversity problems in the population.  We now believe that
the population distribution is to blame in these situations; something
that we will explore in detail in Section~\ref{secSAT}.

\begin{figure*}[t]
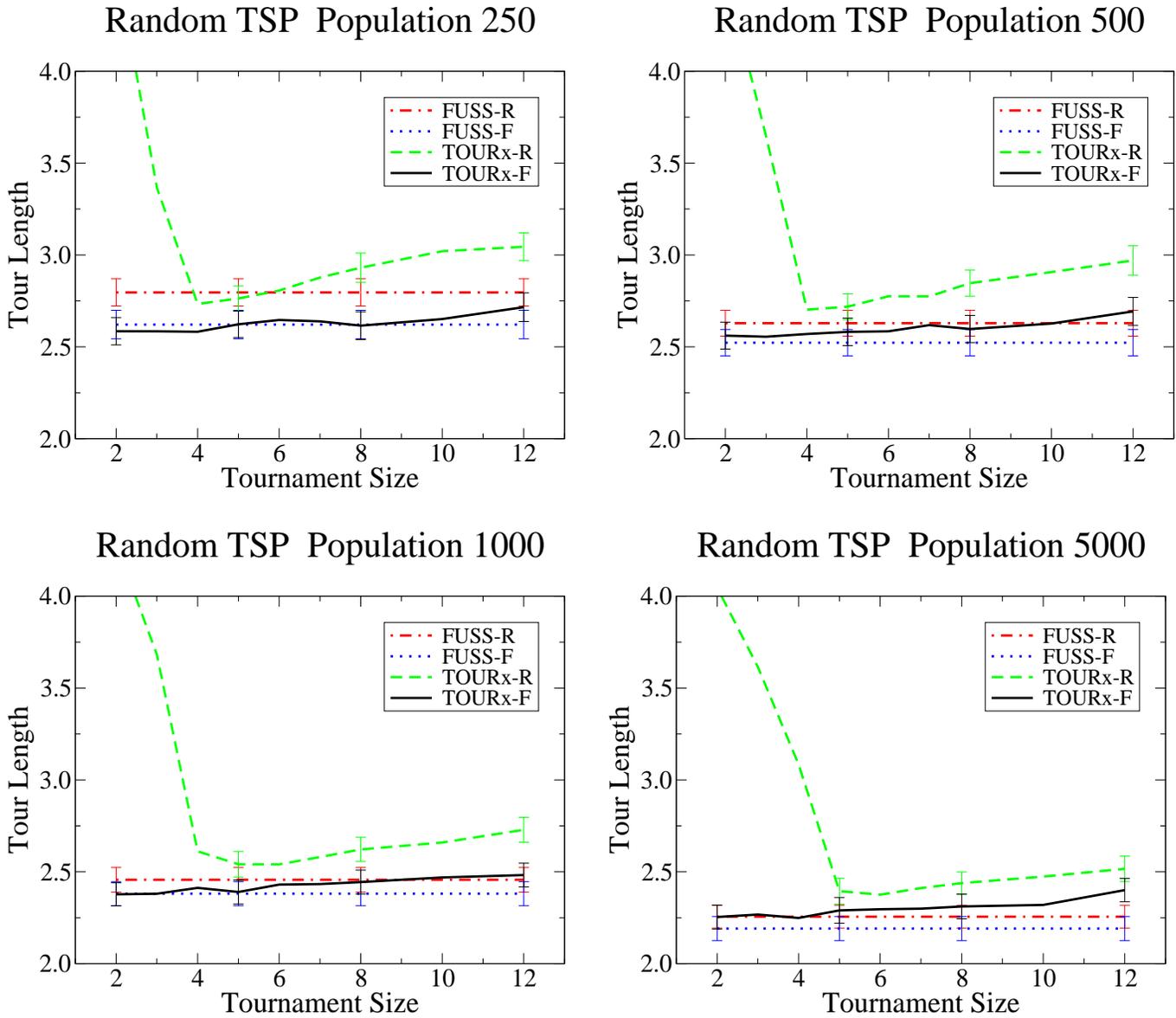

\includegraphics[width=0.48\textwidth,height=0.30\textheight]{dtpsi20-s40-p250.eps}
\includegraphics[width=0.02\textwidth]{space.eps}
\includegraphics[width=0.48\textwidth,height=0.30\textheight]{dtpsi20-s40-p500.eps}
\includegraphics[width=1.00\textwidth,height=0.02\textheight]{space.eps}
\includegraphics[width=0.48\textwidth,height=0.30\textheight]{dtpsi20-s40-p1k.eps}
\includegraphics[width=0.02\textwidth]{space.eps}
\includegraphics[width=0.48\textwidth,height=0.30\textheight]{dtpsi20-s40-p5k.eps}
\caption{\label{tsp} The performance of TOURx-F is much more stable
than TOURx-R under variation in the selection intensity.  Also both
FUSS-R and FUSS-F produce very good results, especially with the
larger populations.}
\end{figure*}

The second graph in Figure~\ref{DTSP-1} shows the same set of
selection schemes but now using FUDS as the deletion scheme.  With
FUDS the performance of all of the selection schemes either stayed the
same or improved.  In the case of TOUR3 the improvement was dramatic
and for TOUR12 the improvement was also quite significant.  This is
interesting because it shows that with fitness uniform deletion,
performance can improve when the selection intensity is either too
high or too low.  That is, when using FUDS the performance of the EA
now appears to be more robust with respect to variation in selection
intensity.

In the case of TOUR12-F this is evidence of improved population
diversity as the EA is no longer becoming stuck.  However for TOUR3-R
the selection intensity is quite low and thus we would expect the
population diversity to be relatively good.  Thus the fact that
TOUR3-F was so much better than TOUR3-R suggests that FUDS can have
significant performance benefits that are not related to improved
population diversity.

Investigating further it seems that this effect is due to the way that
FUDS focuses the deletion on the large mass of individuals which have
an average level of fitness while completely leaving the less common
fit individuals alone.  This helps a system with very weak selection
intensity move the mass of the population up through the fitness
space.  With higher selection intensity this problem tends not to
occur as individuals in this central mass are less likely to be
selected thus reducing the rate at which new individuals of average
fitness are added to the population.

In order to better understand how stable FUDS performance is when used
with different selection intensities we ran another set of tests on
random TSP problems with 20 cities and graphed how performance varied
by tournament size.  For these tests we set the EA to stop each run
when no improvement had occurred in 40 generations.  We also tested on
a range of population sizes: 250, 500, 1000 and 5000.  The results
appear in Figure~\ref{tsp}.

In these graphs we can now clearly see how the performance of TOURx-R
varies significantly with tournament size.  Below the optimal
tournament size performance worsened quickly while above this value it
also worsened, though more slowly.  Interestingly, with a population
size of 5000 the optimal tournament size was about 6, while with small
populations the optimal value fell to just 4.  Presumably this was
partly because smaller populations have lower diversity and thus
cannot withstand as much selection intensity.

In contrast FUSS-R and FUSS-F appear as horizontal lines as they do
not have a tournament size parameter.  We see that they have performed
as well as the optimal performance of TOURx-R without requiring any
tuning.  Indeed for larger populations FUSS-R appears to be even
better than the optimally tuned performance of TOURx-R.  This is a
very positive result for the parameterless FUSS.

Comparing FUDS with random deletion we also see impressive results.
For every combination of selection scheme, tournament size and
population size the result with FUDS was better than the corresponding
result with random deletion, and in some cases much better.
Furthermore these graphs clearly display the improved robustness of
tournament selection with FUDS as TOURx-F produced near optimal
results for all tournament sizes.  Even with an optimally tuned
tournament size FUDS increased performance, particularly with the
smaller populations.  Indeed for each population size tested the worst
performance of TOURx-F was equal to the best performance of
\mbox{TOURx-R}.

With FUSS there was also a performance advantage when using FUDS,
again more so with the smaller populations.  The combination of both
FUSS and FUDS was especially effective as can be seen by the
consistently superior performance of FUSS-F across all of the graphs.

More tests were run exploring performance with up to 100 cities.
Although the performance of FUDS remained stronger than random
deletion for very low selection intensity, for high selection
intensity the two were equal.  We believe that the reason for this is
the following: When the space of potential solutions is very large
finding anything close to a global optimum is practically impossible,
indeed it is difficult to even find the top of a reasonable local
optimum as the space has so many dimensions.  In these situations it
is more important to put effort into simply climbing in the space
rather than spreading out and trying to thoroughly explore.  Thus
higher selection intensity can be an advantage for large problem
spaces.  At any rate, for large problems and with high selection
intensity FUDS did not appear to hinder the performance, while with
low selection intensity it continued to significantly improve it.

\begin{figure*}[t]
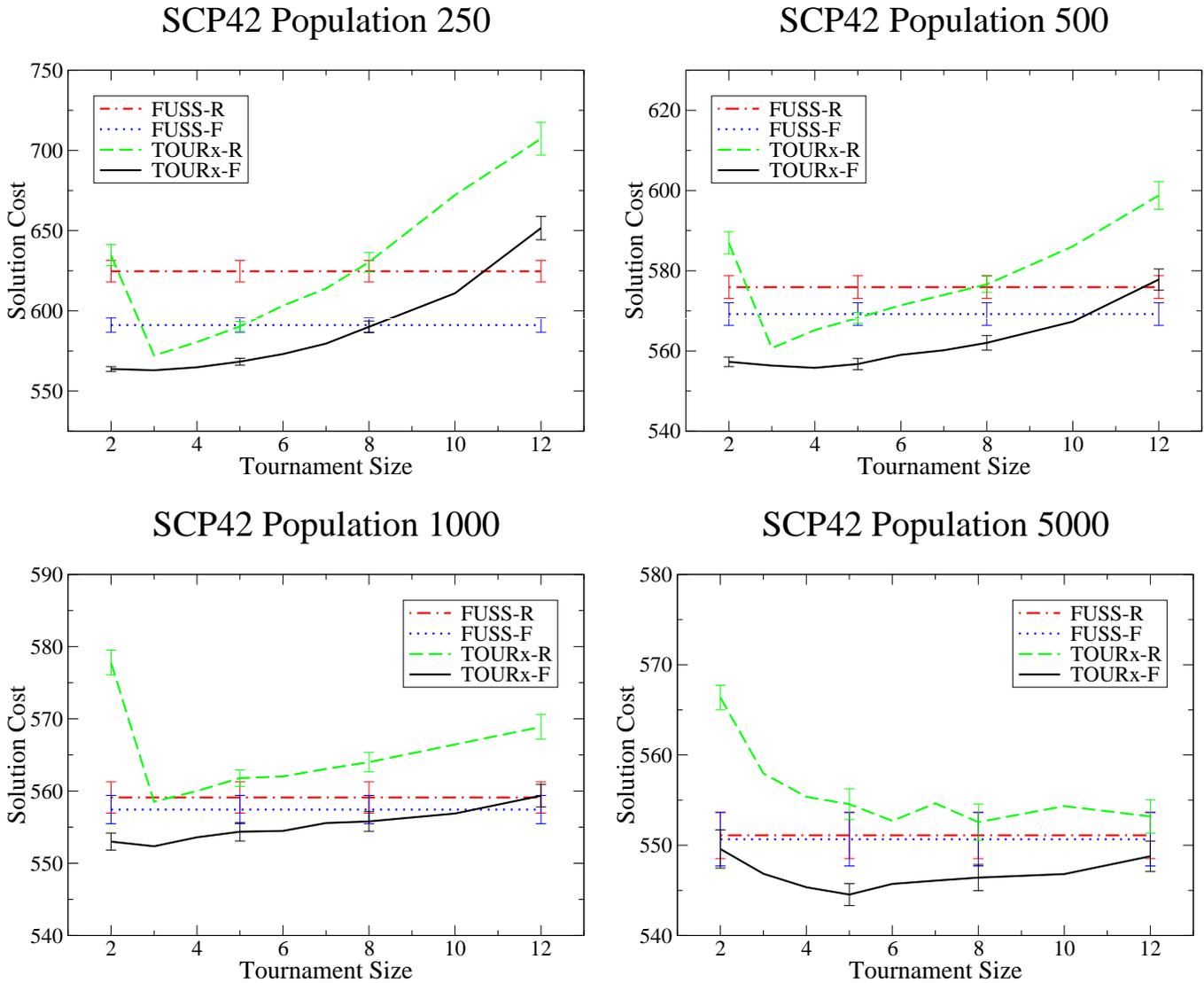

\includegraphics[width=0.485\textwidth]{SCPI42-p250.eps}
\includegraphics[width=0.01\textwidth]{space.eps}
\includegraphics[width=0.485\textwidth]{SCPI42-p500.eps}
\includegraphics[width=1.00\textwidth,height=0.01\textheight]{space.eps}
\includegraphics[width=0.485\textwidth]{SCPI42-p1k.eps}
\includegraphics[width=0.01\textwidth]{space.eps}
\includegraphics[width=0.485\textwidth]{SCPI42-p5k.eps}
\caption{\label{scp-unbal} The performance of FUSS for the two smaller
populations was relatively poor, while for the larger populations it
matched the optimal performance of TOURx-R.  FUDS again produced
superior results to random deletion in all situations tested.}
\end{figure*}

Experiments were also performed using the more efficient ``2-Opt''
mutation operator.  As expected, this increased performance and allowed
much higher selection pressure to be used.  Of course the problem then
no longer had the kind of deceptive structure that heavily punishes
high selection pressure that we are looking for.  Nevertheless, FUDS
continued to significantly boost the performance of tournament
selection, in particular when the tournament size was too small.

\section{Set Covering Problem}\label{secSetCover}

The set covering problem (SCP) is a reasonably well known NP-complete
optimization problem with many real world applications.  Let $M \in
\{0,1\}^{m \times n}$ be a binary valued matrix and let $c_j > 0$ for
$j \in \{1, \ldots n \}$ be the cost of column $j$.  The goal is to
find a subset of the columns such that the cost is minimized.  Define
$x_j = 1$ if column $j$ is in our solution and 0 otherwise.  We can
then express the cost of this solution as $\sum_{j=1}^n c_j x_j$
subject to the condition that $\sum_{j=1}^n m_{ij} x_j \geq 1$ for $i
\in \{1, \ldots m\}$.

Our system of representation, mutation operators and crossover follow
that used by Beasley \cite{Beasley:96} and we compute the fitness by
taking the reciprocal of the cost.  The results presented here are
based on the ``scp42'' problem from a standard collection of SCP
problems \cite{Beasley:03}.  The results obtained on other problems in
this test set were similar.  We found that increasing the crossover
probability and reducing the mutation probability improved
performance, especially when the selection intensity was low.  Thus we
have tested the system with a crossover probability of 0.8 and a
mutation probability of 0.2.  We performed each test at least 50 times
in order to minimize the error bars.  Our stopping criterion was to
terminate each run after no improvement in minimal cost
had occurred
for 40 generations.  The results for this test appear in
Figure~\ref{scp-unbal}.

\begin{figure*}[t]
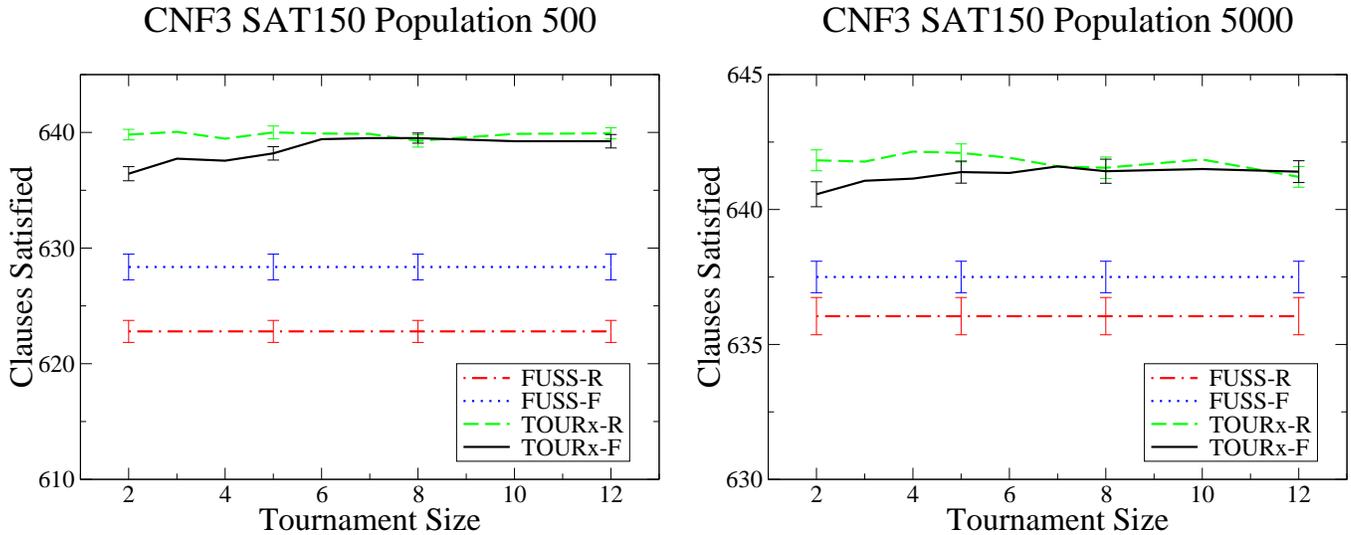

\includegraphics[width=0.485\textwidth]{CNF150-p500.eps}
\includegraphics[width=0.01\textwidth]{space.eps}
\includegraphics[width=0.485\textwidth]{CNF150-p5k.eps}
\caption{\label{cnf-all}With low selection intensity TOURx-F performed
slightly below TOURx-R, but was otherwise comparable.  FUSS had
serious difficulties.}
\end{figure*}

Similar to the TSP graphs we again see the importance of correctly
tuning the tournament size with TOURx-R.  We also see the optimal
range of performance for TOURx-R moving to the right as the population
sizes increases.  This is what we would expect due to the greater
diversity in larger populations.  This kind of variability is one of
the reasons why the selection intensity parameter usually has to be
determined by experimentation.

Unlike with TSP however, the performance of FUSS was less convincing
in these results.  With the smaller populations of 250 and 500 FUSS-R
was only better than TOURx-R when the tournament size was very low or
very high.  With the larger populations of 1,000 and 5,000 the results
were much better with FUSS-R performing as well as the optimal
performance of TOURx-R.  FUSS-F performed better than FUSS-R, in
particular with the smaller populations though this improvement was
still insufficient for it to match the optimal performance of TOURx-R
in these cases.  The fact that the performance of FUSS varied by
population size suggests that FUSS might be experiencing some kind of
population diversity problem.  We will look more carefully at
diversity issues in the next section.

With FUDS the results were again very impressive.  As with the TSP
tests; for all combinations of selection scheme, tournament size and
population size that we tested, the performance with FUDS was superior
to the corresponding performance with random deletion.  This was true
even when the tournament size was optimal.  While the performance of
TOURx-F did vary significantly with different tournament sizes, the
results were more robust than TOURx-R, especially with the larger
populations.  Indeed for the larger two populations we again have a
situation where the worst performance of TOURx-F is equal to the
optimal performance of TOURx-R.

\section{Maximum CNF3 SAT}\label{secSAT}

Maximum CNF3 SAT is a well known NP hard optimization problem
\cite{Crescenzi:04} that has been extensively studied.  A three
literal conjunctive normal form (CNF) logical equation is a boolean
equation that consists of a conjunction of clauses where each clause
contains a disjunction of three literals.  So for example, $(a \lor b
\lor \lnot c) \land ( a \lor \lnot e \lor f)$ is a CNF3 expression.
The goal in the maximum CNF3 SAT problem is to find an instantiation
of the variables such that the maximum number of clauses evaluate to
true.  Thus for the above equation if $a = F$, $b = T$, $c = T$, $e =
T$, and $f = F$ then just one clause evaluates to true and thus this
instantiation gets a score of one.  Achieving significant results in
this area would be difficult and this is not our aim; we are simply
using this problem as a test to compare selection and deletion
schemes.

Our test problems have been taken from the SATLIB collection of SAT
benchmark tests \cite{Hoos:00}.  The first test was performed on the
full set of 100 instances of randomly generated CNF3 formula with 150
variables and 645 clauses, all of which are known to be satisfiable.
Based on test results the crossover and mutation probabilities were
left at the default values.  Our mutation operator simply flips one
boolean variable and the crossover operator forms a new individual by
randomly selecting for each variable which parent's state to take.
Fitness was simply taken to be the number of classes satisfied.  Again
we tested across a range of tournament sizes and population sizes.
The results of these tests appear in Figure~\ref{cnf-all}.

We have shown only the population sizes of 500 and 5,000 as the other
population sizes tested followed the same pattern.  Interestingly for
this problem there was no evidence of better performance with FUDS at
higher selection intensities.  Nor for that matter was there the
decline in performance with TOURx-R that we have seen elsewhere.
Indeed with random deletion the selection intensity appeared to have
no impact on performance at all.  While SAT3 CNF is an NP hard
optimization problem, this lack of dependence of our selection
intensity parameter suggests that it may not have the deceptive
structure that FUSS and FUDS are designed for.

With low selection intensity FUDS caused performance to fall below
that of random deletion; something that we have not seen before.
Because the advantages of FUDS have been more apparent with low
populations in other test problems, we also tested the system with a
population size of only 150. Unfortunately no interesting changes in
behavior were observed.

\begin{figure*}[t]
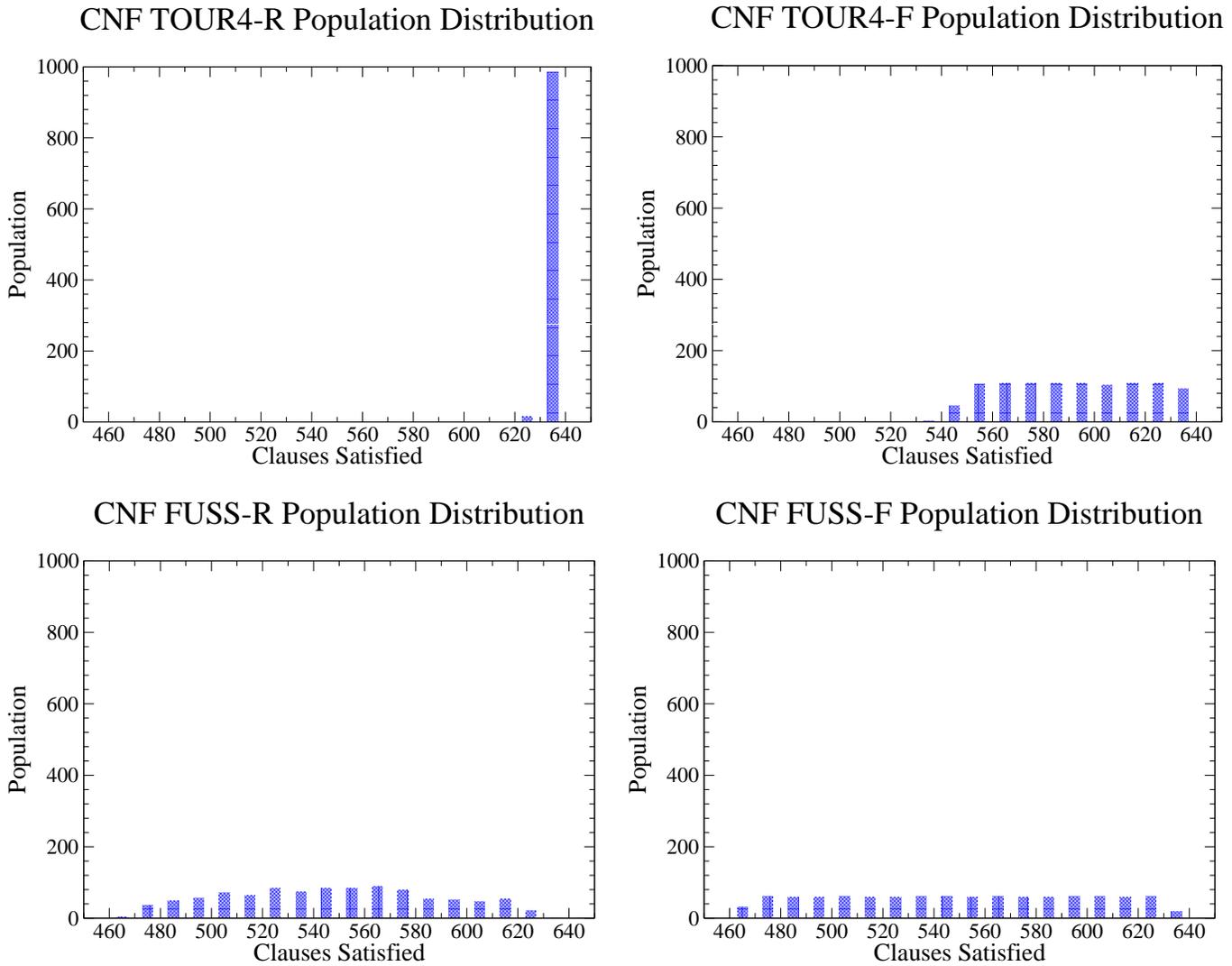

\includegraphics[width=0.485\textwidth]{CNF-TOUR4-R-popDist.eps}
\includegraphics[width=0.01\textwidth]{space.eps}
\includegraphics[width=0.485\textwidth]{CNF-TOUR4-B-popDist.eps}
\includegraphics[width=1.00\textwidth,height=0.01\textheight]{space.eps}
\includegraphics[width=0.485\textwidth]{CNF-FUSS-R-popDist.eps}
\includegraphics[width=0.01\textwidth]{space.eps}
\includegraphics[width=0.485\textwidth]{CNF-FUSS-B-popDist.eps}
\caption{\label{cnf-pop} With TOUR4-R the population collapses to a
narrow band of fitness levels while with TOUR4-F the distribution is
flat.  Under FUSS the population spreads out in both directions with
FUSS-F in particular giving an extremely uniform distribution.}
\end{figure*}

While FUDS had minor difficulties, FUSS had serious problems for all
the population sizes that we tested.  We suspected that the uniform
nature of the population distribution that should occur with both FUSS
and FUDS might be to blame as we only expect this to be a benefit for
very deceptive problems which are sensitive to the tuning of the
selection intensity parameter.  Thus we ran the EA with a population
of 1000 and graphed the population distribution across the number of
clauses satisfied at the end of the run.  We stopped each run when the
EA made no progress in 40 generations.  The results of this appear in
Figure~\ref{cnf-pop}.

The first thing to note is that with TOUR4-R the population collapses
to a narrow band of fitness levels, as expected.  With TOUR4-F the
distribution is now uniform, though practically none of the population
satisfies fewer than 550 clauses.  The reason for this is quite
simple: While FUDS levels the population distribution out, TOUR4 tends
to select the most fit individuals and thus pushes the population to
the right from its starting point.  In contrast, FUSS pushes the
population toward currently unoccupied fitness levels.  This results
in the population spreading out in both directions and so the number
of individuals with extremely poor fitness is much higher.

Given that our goal is to find an instantiation that satisfies all 645
clauses, it is questionable whether having a large percentage of the
population unable to satisfy even 600 clauses is of much benefit.
While the total population diversity under FUSS-F might be very high,
perhaps the kind of diversity that matters the most is the diversity
among the relatively fit individuals in the population.  This should
be true for all but the most excessively deceptive problems.  By
thinly spreading the population across a very wide range of fitness
levels we actually end up with very few individuals with the kind of
diversity that matters.  Of course this depends on the nature of the
problem we are trying to solve and the fitness function that we use.

\begin{figure*}[t]
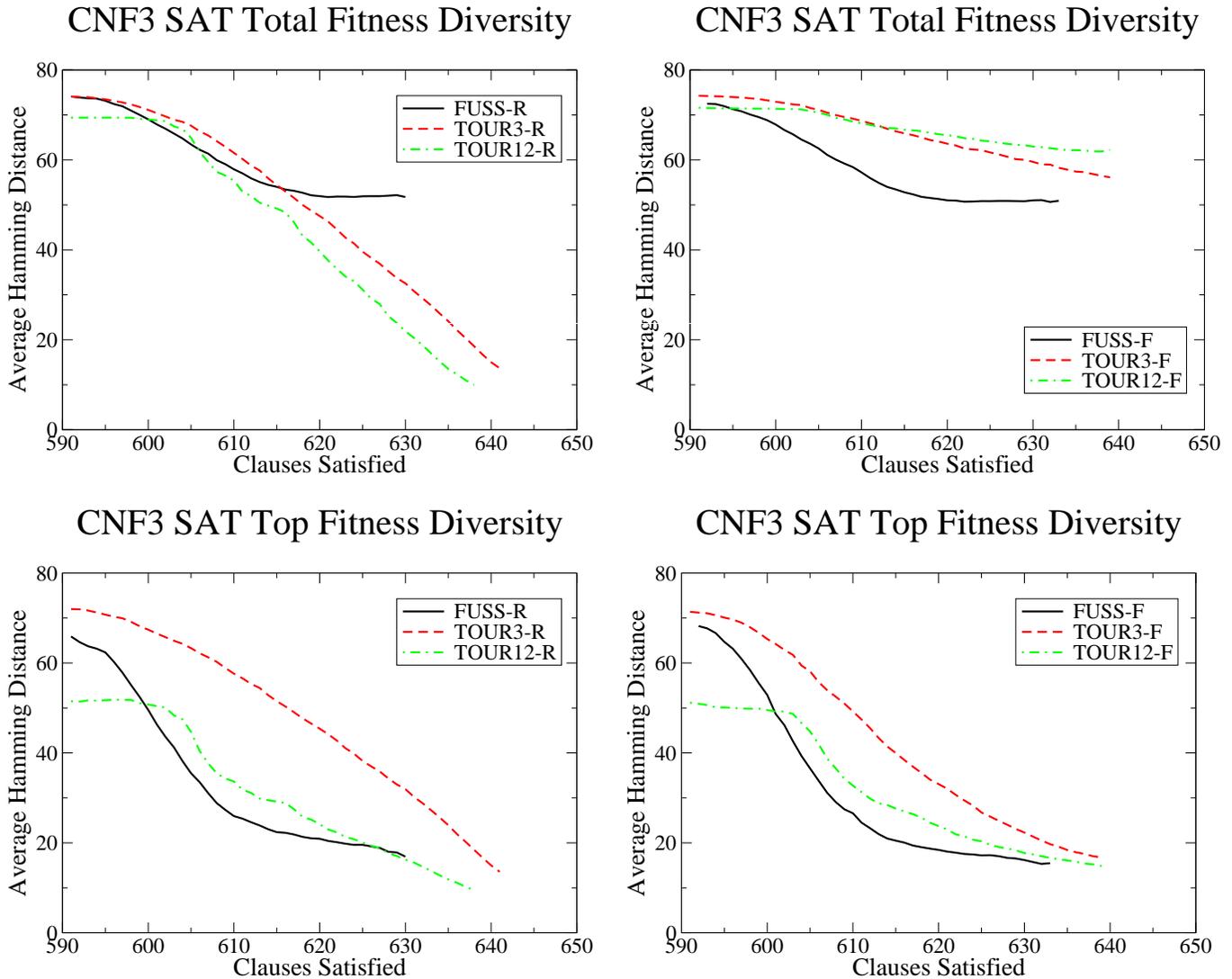

\includegraphics[width=0.485\textwidth]{CNF-total-diversity-R.eps}
\includegraphics[width=0.01\textwidth]{space.eps}
\includegraphics[width=0.485\textwidth]{CNF-total-diversity-F.eps}
\includegraphics[width=1.00\textwidth,height=0.01\textheight]{space.eps}
\includegraphics[width=0.485\textwidth]{CNF-top-diversity-R.eps}
\includegraphics[width=0.01\textwidth]{space.eps}
\includegraphics[width=0.485\textwidth]{CNF-top-diversity-F.eps}
\caption{\label{cnf-diver} While the total population diversity is
very strong under FUSS, the diversity among fit individuals is weak.
FUDS improves the total population diversity compared to random
deletion, but has little effect on the diversity among the fit
individuals.}
\end{figure*}

Fortunately with CNF3 SAT we can directly measure population diversity
by taking the average hamming distance between individuals' genomes.
While this means that the value of the fitness based similarity metric
is questionable for this problem, as more direct methods like crowding
can be applied, it is a useful situation for our analysis as it allows
us to directly measure how effective FUSS and FUDS are at preserving
population diversity.  The hope of course is that any positive
benefits that we have seen here will also carry over to problems where
directly measuring the diversity is problematic.

For the diversity tests we used a population size of 1000 again.  For
comparison we used FUSS, TOUR3 and TOUR12 both with random deletion
and with FUDS.  In each run we calculated two different statistics:
The average hamming distance between individuals in the whole
population, and the average hamming distance between individuals whose
fitness was no more than 20 below the fittest individual in the
population at the time.  These two measurements give us the ``total
population diversity'' and ``high fitness diversity'' graphs in
Figure~\ref{cnf-diver}.

We graphed these measurements against the solution cost of the fittest
individual rather than the number of generations.  This is only fair
because if good solutions are found very quickly then an equally rapid
decline in diversity is acceptable and to be expected.  Indeed it is
trivial to come up with a system which always maintains high
population diversity how ever long it runs, but is unlikely to find
any good solutions.  The results were averaged over all 100 problems in
the test set.  Because the best solution found in each run varied we
have only graphed each curve until such a point where fewer than 50\%
of the runs were able to achieve this level of fitness.  Thus the
terminal point at the right of each curve is representative of fairly
typical runs rather than just a few exceptional ones that perhaps
found unusually good solutions by chance.

The top two graphs in Figure~\ref{cnf-diver} show the total population
diversity.  As expected the diversity with TOUR3-R and TOUR12-R
decline steadily as finding better solutions becomes increasingly
difficult and the population tends to collapse into a narrow band of
fitness.  As we would expect, the total population diversity with
TOUR3-R is higher than with TOUR12-R.  While FUSS-R declines initially
it then stabilizes at around 50 before becoming stuck.  As the TOUR3-R
and TOUR12-R curves both extend further to the right, even though the
total population diversity becomes quite low, this show that diversity
problems in the population as a whole are not a significant factor
behind the performance problems with FUSS-R.

The top right graph shows the same selection schemes, but this time
with FUDS.  As expected FUDS has significantly improved the total
population diversity with both TOUR3 and TOUR12, while having little
impact on FUSS which already has a relatively flat population
distribution.  As the maximal solution found by TOUR3-F and TOUR12-F
were not better than TOUR3-R and TOUR12-R, this indicates that
improved total population diversity is not a significant factor in the
performance of the EA for this type of optimization problem.  That
FUDS has lifted the total diversity for TOUR3 and TOUR12 so that they
are now above FUSS-F, is particularly interesting.  This suggests that
while FUSS has high total population diversity, there appears to be
some more subtle effects that are causing the diversity to be lower
than it could be.  It may be related to the fact the FUSS sometimes
heavily selects from small groups within the population during the
early stages of the optimization process, as we noted in
Section~\ref{secTSP}.  However we are not certain whether this is
occurring in this case.

On the lower set of graphs we see the diversity among the fitter
individuals in the population; specifically those whose fitness is no
more than 20 below the fittest individual in the population at the
time.  On the first graph on the left we see that TOUR3 has
significantly greater diversity than TOUR12 with both deletion
schemes.  This is expected as TOUR3 tends to search more evolutionary
paths while TOUR12 just rushes down a few.  Disappointingly FUDS does
not appear to have made much difference to the diversity among these
highly fit individuals, though the curves do flatten out a little as
the diversity drops below 30, so perhaps FUDS is having a slight
impact.

For both FUSS-R and FUSS-F the diversity among the fit individuals was
poor, indeed it was even worse than TOUR12 for both deletion schemes.
Thus, while the total population diversity with FUSS tends to be high,
the diversity among the fittest individuals in the population can be
quite poor.  Furthermore, the curves for high fitness diversity all
end once the diversity drops into the 12 to 17 range.  As this pattern
was absent from the graphs of total population diversity, this
indicates that it is indeed the diversity among the relatively fit
individuals in the population that most determines when the EA is
going to become stuck.

In summary, these results show that while FUSS has been successful in
maximizing total population diversity, for problems such as CNF3 SAT
this is not sufficient.  It appears to be more important that the EA
maximizes the diversity among those individuals which have higher
fitness and in this regard FUSS is poor, which leads to poor
performance.  This is most likely a characteristic of optimization
problems which, while still difficult, are not as deceptive as SCP or
random TSP.

\section{Conclusions and Future Research Directions}\label{secConc}

We have addressed the problem of balancing the selection intensity
in EAs, which determines speed versus quality of a solution. We
invented a new fitness uniform selection scheme FUSS. It generates
a selection pressure toward sparsely populated fitness levels.
This property is unique to FUSS as compared to other selection
schemes (STD).
It results in the desired high selection pressure toward higher
fitness if there are only a few fit individuals. The selection
pressure is automatically reduced when the number of fit
individuals increases.
We motivated FUSS as a scheme which bounds the number of {\em
similar} individuals in a population. We defined a universal
similarity relation solely depending on the fitness, independent
of the problem structure, representation and EA details.
We showed analytically by way of a simple example that FUSS can be
much more effective than STD. A joint pair selection scheme for
recombination has been defined.
A heuristic worst case analysis of FUSS compared to STD has been
given. For this, the fitness tree model has been defined, which is an
interesting analytic tool in itself.
FUSS solves the problem of population takeover and the resulting
loss of genetic diversity of STD, while still generating enough
selection pressure. It does not help in getting a more uniform
distribution within a fitness level.

We have also invented a related system called FUDS which achieves a
similar effect to FUSS except that it works through deletion rather
than through selection.  This means that FUDS shares many of the
important characteristics of FUSS including strong total population
diversity and the impossibility of population collapse.  We showed
analytically that for a simple deceptive optimization problem the
performance of STD when used with FUDS scales similarly to FUSS.

A test system has been constructed and used to evaluate the empirical
performance of both FUSS and FUDS on a range of optimization problems
with different population sizes, mutation probabilities and crossover
probabilities.  Their performance has been compared to the more
standard methods of tournament selection and random deletion.  For the
artificial deceptive 2D optimization problem and random distance
matrix TSP problems both FUSS and FUDS performed extremely well.  For
the deceptive 2D problem they dramatically improved the scaling
exponent in the number of generations needed to find the global
optimum.  For the TSP problems FUSS-R performed as well as optimally
tuned TOURx-R for all population sizes, and FUDS caused TOURx to
perform near optimally for all tournament sizes and population sizes.

With SCP problems with small populations the performance of FUSS-R was
only better than TOURx-R when the tournament size was poorly set.  For
populations larger than 1,000 however, FUSS-R continued to perform as
well as the optimal results for TOURx-R.  FUDS was again consistently
superior returning better results than random deletion for every
combination of selection scheme, tournament size and population size
tested.

For CNF3 SAT problems we ran into difficulties however.  While FUDS
significantly improved the performance of FUSS, it was inferior to
random deletion for low selection intensities.  In other cases the
performance was comparable.  FUSS however had serious performance
problems.  Further investigations revealed that this appears to be due
to the small number of individuals in the population that have
relatively high fitness when using FUSS.  We measured the diversity in
the population and found that while the total population diversity
with FUSS was high, the diversity among the fit individuals was
relatively poor.  This produced a serious diversity problem in the
population when combined with the fact that there are relatively few
individuals of high fitness when using FUSS.

As the performance of TOURx-R was not impacted by high selection
intensity on the CNF3 SAT problem this indicates that this problem
does not have the kind of deceptive nature that harshly punishes
greedy exploration that we were looking for.  Perhaps for such
problems a less extreme approach is called for.  For example, rather
than trying to spread the population across all fitness levels
uniformly we should instead control the distribution so that it is
biased toward high fitness but never collapses totally as it does with
TOURx-R.

We have experimented with a deletion scheme which deletes the
population distribution down to a convex curve peaked at the fittest
individual in the population.  This is the deletion equivalent of the
scale independent selection scheme described in
Section~\ref{secCross}.  Our results thus far indicate that the
performance is equal or slightly superior to random deletion in all
situations.  However the dramatic improvements that FUDS has over
random deletion in some cases are now less significant.

Another possibility is to manipulate the fitness function to
effectively achieve the same thing.  For example, we have found that
by taking the fitness to be the reciprocal of the number of
unsatisfied clauses in the CNF3 SAT problem the performance of FUSS
improves significantly, indeed it is then comparable to TOURx.
Perhaps however it would be better to avoid these performance tricks
and instead focus on extremely deceptive problems where high selection
intensity is heavily punished, that is, the kinds of problems that
FUSS and FUDS were specifically designed for.

\subsection{Acknowledgments}
This work was supported by SNF grants 2100-67712.02 and 200020-107616.


\begin{small}

\end{small}

\vspace{5ex}
\begin{wrapfigure}[9]{l}{0.3\columnwidth}
\vspace{-2.5ex}\includegraphics[width=0.3\columnwidth]{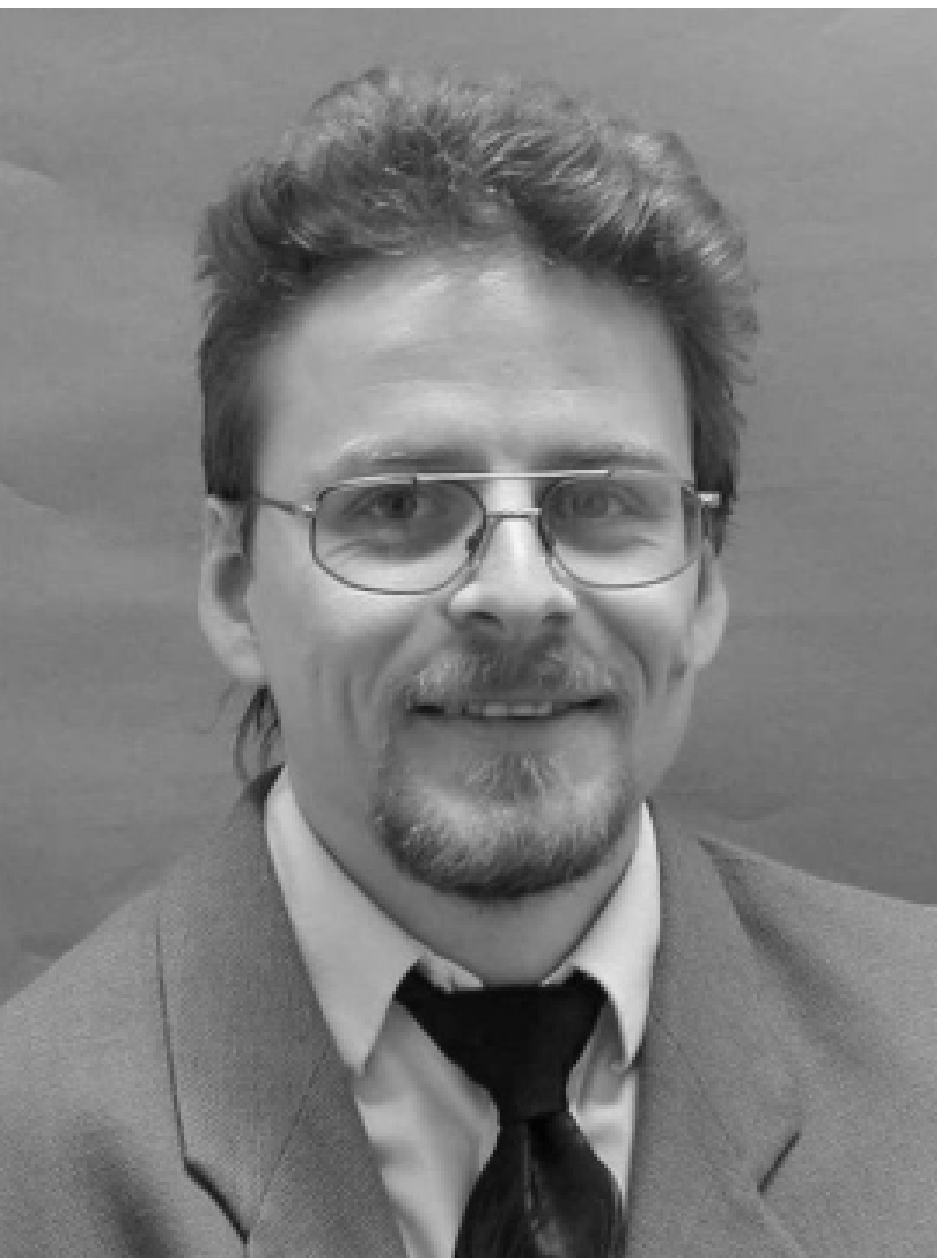}
\end{wrapfigure}
\noindent{\bf Marcus Hutter}
received the M.Sc.\ degree in computer science and the
Ph.D.\ degree in theoretical particle physics from the (Technical)
University, Munich, Germany, in 1992 and 1995, respectively.

Thereafter, he developed algorithms in a medical software company
for five years. Since 2000, he has published over 35 research papers
while a Researcher with the Dalle Molle Institute for Artificial
Intelligence (IDSIA), Lugano, Switzerland. He is the
author of {\em Universal Artificial Intelligence} (EATCS: Springer,
2004). His current interests are centered around reinforcement
learning, algorithmic information theory and statistics, universal
induction schemes, adaptive control theory, and related areas.

\vspace{5ex}
\begin{wrapfigure}[9]{l}{0.3\columnwidth}
\vspace{-2.5ex}\includegraphics[width=0.3\columnwidth]{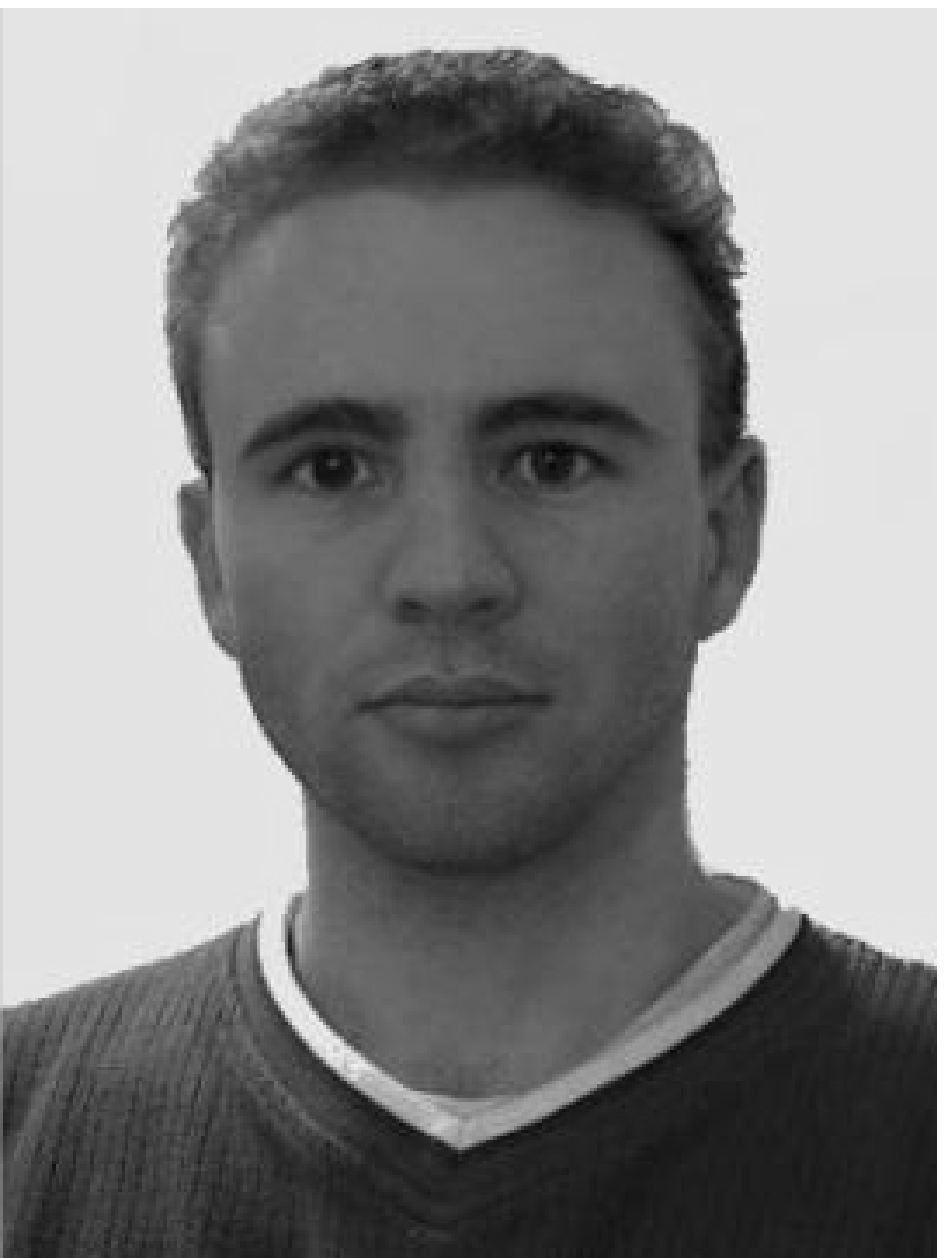}
\end{wrapfigure}
\noindent{\bf Shane Legg}
received the B.C.M.S.\ degree in mathematical and computer sciences
from the University of Waikato, Hamilton, New Zealand, in 1996 and
the M.Sc.\ degree in mathematics from Auckland University, Auckland,
New Zealand, in 1997. He is currently working towards the Ph.D.\
degree at the Dalle Molle Institute for Artificial Intelligence
(IDSIA), Lugano, Switzerland.

After receiving the M.Sc.\ degree in 1997, he then worked in a
number of companies in New Zealand and the United States mainly
focusing on commercial applications of artificial intelligence. His
research is focused on genetic algorithms, complexity theory, and
theoretical models of artificial intelligence.

\end{document}